\documentclass{article}

\usepackage{arxiv}

\usepackage[utf8]{inputenc} 
\usepackage[T1]{fontenc}    
\usepackage{hyperref}       
\usepackage{url}            
\usepackage{booktabs}       
\usepackage{amsfonts}       
\usepackage{nicefrac}       
\usepackage{microtype}      
\usepackage{lipsum}		
\usepackage{graphicx}
\usepackage{natbib}
\usepackage{doi}

\usepackage{mathtools}
\usepackage{amssymb}
\usepackage{algpseudocode}
\usepackage{caption}
\usepackage{subcaption}
\usepackage{multicol}
\usepackage{tablefootnote}
\usepackage{wrapfig}
\usepackage{textcomp}
\usepackage{siunitx}
\usepackage{multirow}
\usepackage{graphics}
\usepackage{multibib}
\usepackage{algorithm}
\usepackage{arydshln}
\usepackage{pifont}
\newcommand{\cmark}{\ding{51}}%
\newcommand{\xmark}{\ding{55}}%
\usepackage[leftcaption]{sidecap}
\usepackage{textcomp}

\title{Learning Decomposable and Debiased Representations via Attribute-Centric Information Bottlenecks}

\date{} 					

\author{ \href{https://orcid.org/0000-0003-4429-3311}{\includegraphics[scale=0.06]{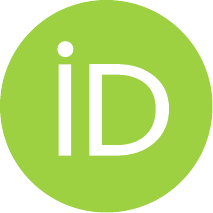}\hspace{1mm}Jinyung Hong}\\
	School of Computing and Augmented Intelligence\\
	Arizona State University\\
	Tempe, AZ 85281 \\
	\texttt{jhong53@asu.edu} \\
	\And
	\href{https://orcid.org/0000-0002-1112-4653}{\includegraphics[scale=0.06]{orcid.pdf}\hspace{1mm}Eun Som Jeon} \\
	School of Arts, Media and Engineering \\
	Arizona State University \\
    Tempe, AZ 85281 \\
	\texttt{ejeon6@asu.edu} \\
    \And
	\href{https://orcid.org/0009-0000-5850-6483}{\includegraphics[scale=0.06]{orcid.pdf}\hspace{1mm}Changhoon Kim} \\
	School of Computing and Augmented Intelligence \\
	Arizona State University \\
    Tempe, AZ 85281 \\
	\texttt{kch@asu.edu} \\
    \And
	\href{https://orcid.org/0009-0004-2912-7248}{\includegraphics[scale=0.06]{orcid.pdf}\hspace{1mm}Keun Hee Park} \\
	School of Computing and Augmented Intelligence \\
	Arizona State University \\
    Tempe, AZ 85281 \\
	\texttt{kpark53@asu.edu} \\
    \And
	\href{https://orcid.org/0009-0001-2546-8190}{\includegraphics[scale=0.06]{orcid.pdf}\hspace{1mm}Utkarsh Nath} \\
	School of Computing and Augmented Intelligence \\
	Arizona State University \\
    Tempe, AZ 85281 \\
	\texttt{unath@asu.edu} \\
    \And
	\href{https://orcid.org/0000-0003-0126-8976}{\includegraphics[scale=0.06]{orcid.pdf}\hspace{1mm}Yezhou Yang} \\
	School of Computing and Augmented Intelligence \\
	Arizona State University \\
    Tempe, AZ 85281 \\
	\texttt{yz.yang@asu.edu} \\
    \And
	\href{https://orcid.org/0000-0002-5263-5943}{\includegraphics[scale=0.06]{orcid.pdf}\hspace{1mm}Pavan Turaga} \\
	School of Arts, Media and Engineering \\
	Arizona State University \\
    Tempe, AZ 85281 \\
	\texttt{pturaga@asu.edu} \\
    \And
	\href{https://orcid.org/0000-0002-7073-6932}{\includegraphics[scale=0.06]{orcid.pdf}\hspace{1mm}Theodore P. Pavlic} \\
	School of Computing and Augmented Intelligence \\
    School of Life Sciences \\
	Arizona State University \\
    Tempe, AZ 85281 \\
	\texttt{tpavlic@asu.edu} \\
}

\renewcommand{\shorttitle}{\textit{arXiv} Template}

\hypersetup{
pdftitle={A template for the arxiv style},
pdfsubject={q-bio.NC, q-bio.QM},
pdfauthor={David S.~Hippocampus, Elias D.~Striatum},
pdfkeywords={First keyword, Second keyword, More},
}

\begin{document}
\maketitle

\begin{abstract}
Biased attributes, spuriously correlated with target labels in a dataset, can problematically lead to neural networks that learn improper shortcuts for classifications and limit their capabilities for out-of-distribution~(OOD) generalization. Although many debiasing approaches have been proposed to ensure correct predictions from biased datasets, few studies have considered learning latent embedding consisting of intrinsic and biased attributes that contribute to improved performance and explain how the model pays attention to attributes. In this paper, we propose a novel debiasing framework, \emph{Debiasing Global Workspace}, introducing attention-based information bottlenecks for learning compositional representations of attributes without defining specific bias types. Based on our observation that learning shape-centric representation helps robust performance on OOD datasets, we adopt those abilities to learn robust and generalizable representations of decomposable latent embeddings corresponding to intrinsic and biasing attributes. We conduct comprehensive evaluations on biased datasets, along with both quantitative and qualitative analyses, to showcase our approach's efficacy in attribute-centric representation learning and its ability to differentiate between intrinsic and bias-related features.
\end{abstract}


\section{Introduction}
\label{sec:intro}
Deep Neural Networks~(DNNs) have achieved remarkable advancements across a spectrum of domains, such as image classification~\cite{he2019bag, xie2020self}, generation~\cite{wang2016generative, kataoka2016image}, and segmentation~\cite{luo2017deep, zheng2014dense}. Despite their impressive achievements, DNNs exhibit vulnerability to dataset bias~\cite{torralba2011unbiased} in scenarios with strong correlations between peripheral attributes and labels within a dataset, and this limits their capabilities for out-of-distribution~(OOD) generalization; models learning shortcuts for classification~\cite{geirhos2020shortcut} and fail to generalize on images with no such correlations during the test phase.

Bias can manifest itself in the presence of a high number of \emph{bias-aligned} samples where labels are correlated with robust and biased features as well as the presence of \emph{bias-conflicting} samples that overrepresent rare features for a label. Because models trained on biased datasets face the issue of strongly favoring bias-aligned samples, many debiased learning approaches allow models to ignore biased features. However, identifying biased features from independent and identically distributed~(i.i.d.) data is complex and requires additional assumptions or supervision from non-i.i.d. training samples~\cite{scholkopf2021toward}.

Some debiasing techniques are based on the assumption that biased features are ``easier to learn'' than robust ones~\cite{scimeca2021shortcut, shah2020pitfalls} and consequently train two models---an auxiliary model that intentionally relies on biased features and a desired debiased model. In these cases, the training process for the debiased model is guided by the auxiliary one~\cite{nam2020learning, sanh2020learning}. For example, training the auxiliary model with a generalized cross-entropy~(GCE) loss~\cite{zhang2018generalized} helps to improve robustness on biased, easy-to-learn features~\cite{nam2020learning}. However, the debiased model is trained by either re-weighting samples~\cite{liu2021just, nam2020learning} or data augmentation~\cite{kim2021biaswap, lee2021learning}. Re-weighting-based methods are straightforward, but they may only be effective when more bias-conflicting examples exist. Furthermore, augmentation-based methods rely on complex generative models or disentangled representations which are challenging to apply to real-world data.

In this paper, we propose \emph{Debiasing Global Workspace}, a novel debiasing approach for attribute-centric representation learning using attention-based information bottlenecks, which can explain how the model pays attention to attributes in a given dataset. 
In our method, the latent embeddings composing attributes, \emph{concepts}, can be learned and those concepts are used to improve the intermediate representations of intrinsic and biased attributes. First, we present our observations showing that attention-based information bottlenecks in explainable AIs can learn strong shape-centric representation and contribute to achieving robust OOD performance compared to other DNNs and that our debiasing method is based on it~(Section~\ref{sec:concept_learning}). Then, we describe the details of our debiasing method along with our training schemes~(Section~\ref{sec:method}). Finally, we evaluate our method on standard debiasing benchmarks and provide comprehensive quantitative and qualitative analyses to showcase our approach's interpretability in capturing attribute representation and clustering performance~(Section~\ref{sec:exp}).

\section{Importance of Interpretable Concept Learning}
\label{sec:concept_learning}
\begin{wrapfigure}{r}{0.35\textwidth}
\vspace{-2.5em}
	\includegraphics[width=0.95\linewidth]{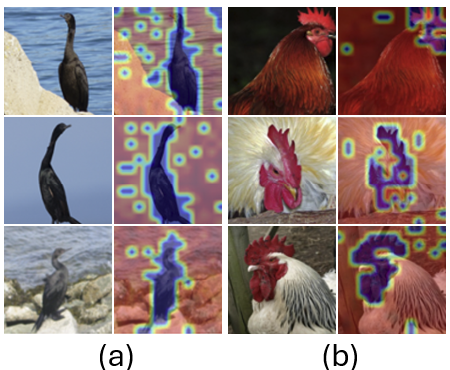}
	\caption{CCT Activation Visualizations. (a) illustrates CCT's identification of bird body contours. (b) highlights the hen's crest contours. Both demonstrate CCT's concept focus without ground truth concept data.}
 \label{fig:cct_vis} 
\vspace{-2em}
\end{wrapfigure}
In this section, we validate that interpretable ML methods can perform shape-centric representation learning through concept learning. Furthermore, we describe the details of experiments to observe whether they can achieve robust performance against various image distortions, demonstrating their applicability in debiasing approaches. 
Additional details are described in Appendix~\ref{app:analysis}.

\subsection{Intrinsically Interpretable Models in XAI}
\label{subsec:xai_cct}
Explainable AI~(XAI) approaches seek to improve the transparency of model decision-making. Generally speaking, models can be analyzed post hoc to extract features that seem to be central to the model's decision-making. However, additional computations are required to extract these explanations, and the resulting explanations typically have a narrow focus on low-level features~\cite{kim2018interpretability, alvarez2018towards, kindermans2019reliability, su2019one}. Alternatively, models can be specially structured to maximize their natural interpretability~\cite{rudin2019stop}. Such \emph{intrinsically interpretable models} promise to be easily understandable by humans by basing decisions on ``concepts'' akin to the foundation of domain expertise~\cite{alvarez2018towards, barbiero2022entropy, chen2019looks, chen2020concept, ghorbani2019towards, goyal2019explaining, kazhdan2020now, kim2018interpretability, koh2020concept, li2018deep, vaswani2017attention, yeh2020completeness, zarlenga2022concept, Hong_2024_WACV}.

\paragraph{Concept-Centric Transformers.}

We focus on the intrinsically interpretable Concept-Centric Transformers (CCTs)~\cite{Hong_2024_WACV}, which can determine how much each image feature contributes to each concept using an attention mechanism~\cite{vaswani2017attention} and provide meaningful concept embedding representations. CCT can identify and highlight semantically significant concepts of images without using human-annotated concept information. 

As shown in Fig.~\ref{fig:cct_vis}, our empirical evaluation validates the capabilities of CCT in identifying and discerning the essential concepts of images. Notably, CCT autonomously identifies and underscores semantically significant concepts in images—accomplished without the aid of human-annotated ground-truth concept data. For instance, in Fig.~\ref{fig:cct_vis}(a), CCT distinctively highlights the contours of a bird's body, pinpointing this feature as a key visual concept. In addition, Fig.~\ref{fig:cct_vis}(b) displays CCT's adeptness in isolating and emphasizing the contour of a hen's crest. These capabilities increase image recognition performance and represent the potential to reduce bias by focusing on crucial visual concepts. Additional implementation details can be found in Appendix~\ref{subapp:implementation_detail}.

\subsection{Analysis on Out-of-Distribution Datasets}
\label{subsec:ood_evaluation}

\paragraph{Our Hypothesis.}
Convolutional Neural Networks~(CNNs) are one of the most famous machine learning models that learn the features used for object recognition rather than manually designed. In~\cite{geirhos2018imagenet}, the widespread belief that CNNs use increasingly complex shape features to recognize objects is incorrect; in fact, it has been found that CNNs place more emphasis on the texture of images, and this behavior resulted in significant performance degradation for distorted images. Furthermore, it has been shown that training models using specialized images with style transfer can overcome this by learning shape-centric representation~\cite{geirhos2018imagenet, singh2023improving}. 
We hypothesize that biased attributes in biased datasets can be seen as one of the texture-bias issues contributing to misleading model training.

Some debiasing approaches~\cite{li2020shape, geirhos2018imagenet} have been proposed considering CNNs' biased behavior on textures, but those were either trained on specialized datasets or used predefined bias types.
In contrast, we argue that shape-centric representation learning methods can be a novel and more fundamental alternative. As we confirmed above, CCT implicitly captures the semantic shapes of concepts well; we conduct quantitative analysis based on the experiment setup in~\cite{geirhos2018imagenet, geirhos2021partial} that allows for quantifying texture and shape biases in CCT and demonstrate how well CCT's ability to learn shape-centric representations contributes to robust performance on diverse out-of-distribution~(OOD) datasets. 
We trained CCT on ImageNet-1K~\cite{deng2009imagenet} following~\cite{Hong_2024_WACV}, and evaluated the model on various OOD datasets in~\cite{geirhos2018imagenet}. 
\begin{figure*}[t!]
     \centering
     \begin{subfigure}[b]{0.48\textwidth}
         \centering
         \includegraphics[width=\textwidth]{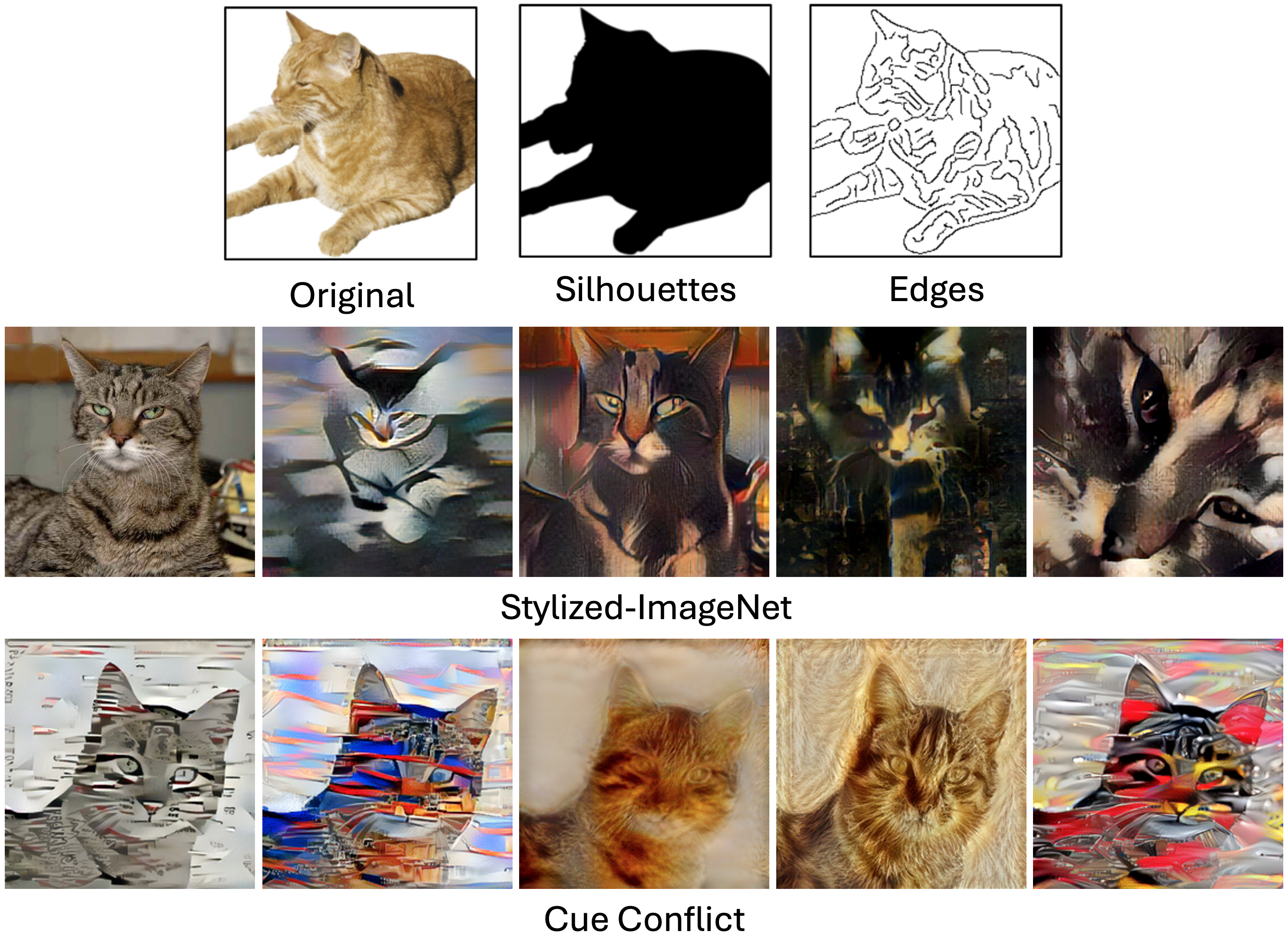}
         \caption{Dataset examples for Original, Silhouettes, Edges, Stylized-ImageNet, and Cue conflict}
         \label{fig:ood_datasets}
     \end{subfigure}
     \hspace{20pt}
     \begin{subfigure}[b]{0.44\textwidth}
         \centering
         \includegraphics[width=\textwidth]{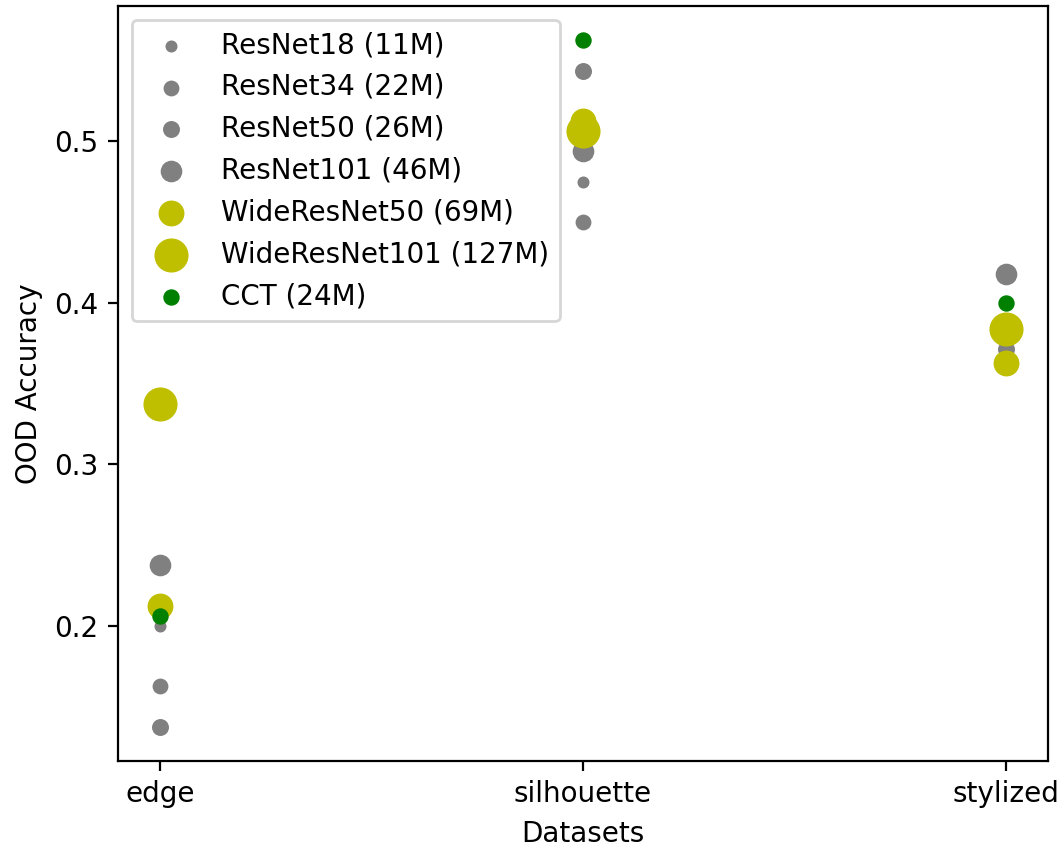}
         \caption{OOD Accuracy Comparison between CCT and other CNN-based models.}
         \label{fig:ood_accuracy}
     \end{subfigure}
    \caption{Out-of-distribution (OOD) Benchmarking.}
    \label{fig:ood_benchmark}
\vspace{-1em}
\end{figure*}

\paragraph{OOD Datasets.}
In OOD generalization test, we utilize a variety of specially prepared datasets in~\cite{geirhos2018imagenet}. These datasets include: \emph{Stylized-ImageNet}, \emph{Edges}, \emph{Silhouettes} and \emph{Cue conflict}.
\textit{Original} consists of 160 natural color images with objects across 10 categories on a white background, \textit{Silhouette} converts images to black silhouettes on a white background;\textit{Edges} converts images to edge-based outlines using Canny edge detector;
\textit{Cue conﬂict} : Generates images by merging the texture from one image with the shape from another through iterative style transfer, creating 1280 images to examine models' ability to resolve visual cue conflicts, and; \textit{Stylized-ImageNet} : Alters ImageNet images by substituting original textures with artistic styles via AdaIN style transfer~\cite{huang2017arbitrary}, focusing analysis on shape rather than texture.
Examples of each dataset are illustrated in Fig.~\ref{fig:ood_benchmark}(a). Check the details of dataset statistics in~\cite{geirhos2018imagenet}.

\paragraph{Experimental Results on OOD Datasets.}
\begin{SCfigure}[0.38][t!]
    \centering
    \includegraphics[width=0.6\textwidth]{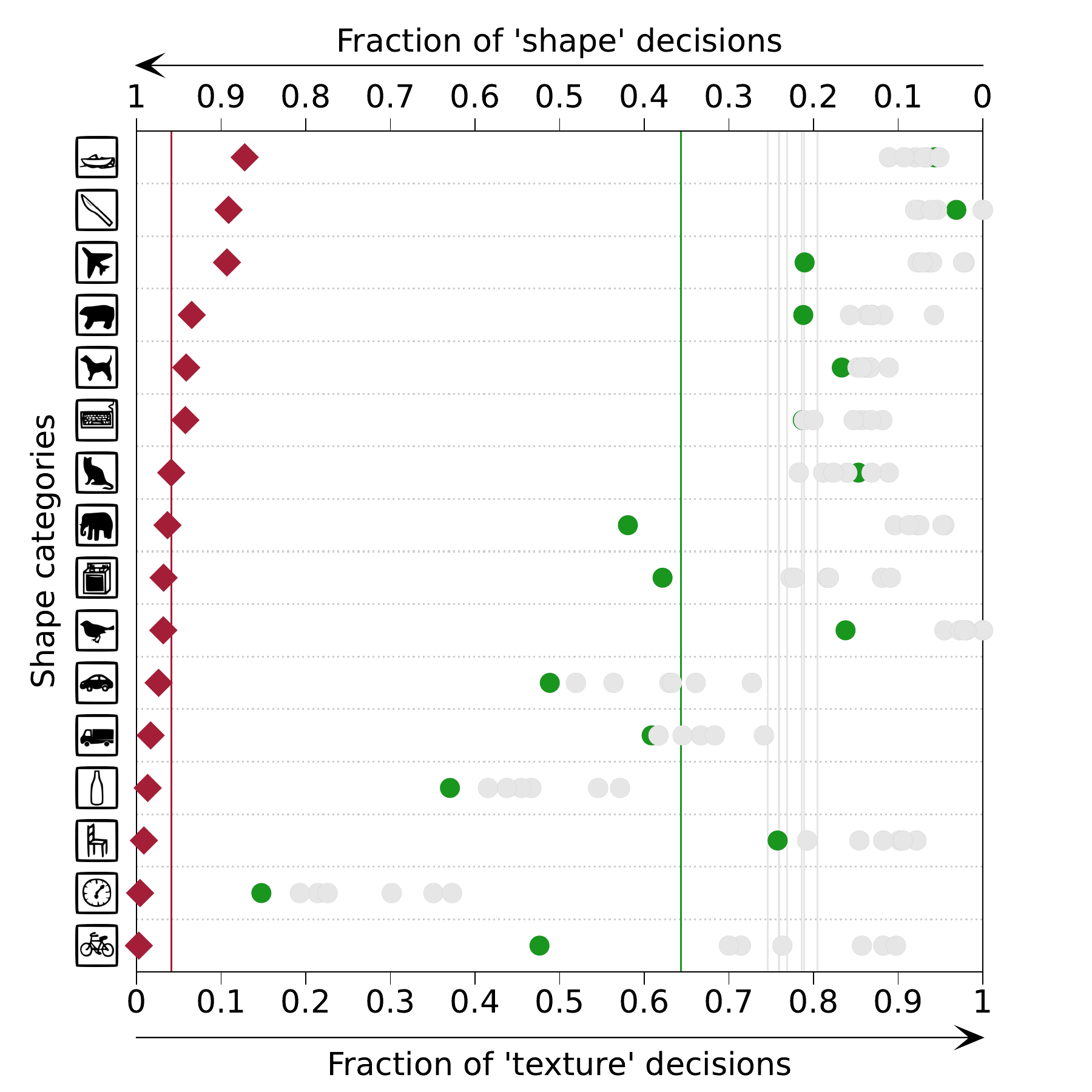}
    \caption{Visualization of classification preferences between shape and texture in images with conflicting cues. Human observers are indicated by red diamonds, various ImageNet-trained CNNs, including multiple ResNet architectures, by grey dots, and the CCT module by green dots. The CCT's results, represented by green dots, demonstrate a higher tendency to prioritize shape over texture, outperforming both standard and more complex ResNet variants, highlighting CCT's efficacy in shape-based recognition despite its relative parameter efficiency. The format of the figure is generated by~\cite{geirhos2021partial}.}
    \label{fig:cue_conflict_result}
\end{SCfigure}
The OOD accuracy presented in Fig.~\ref{fig:ood_benchmark}(b) compares CCT against various ResNet architectures \cite{he2016deep} across Silhouette, stylized ImageNet, and Edge datasets. The results indicate that CCT, represented by dark-green dots, demonstrates superior performance on the Silhouette dataset, achieving the highest accuracy. Even on the Edge and stylized datasets, CCT's efficiency becomes evident when considering its parameter size—24M compared to WideResNet101's 127M, for example, yet still maintaining competitive accuracy. Complementing these findings, Fig.~\ref{fig:cue_conflict_result} underscores CCT's adeptness in learning shape-centric representations, outshining ResNet variants when tested on the Cue Conflict dataset. This proficiency in discerning shapes is pivotal for models aimed at reducing bias, as it relies less on potentially misleading texture cues.

These outcomes collectively affirm that CCT excels in shape-based decision-making, a trait that significantly contributes to its potential for integration into debiasing frameworks. The clear advantage of CCT in handling shape information, even with fewer parameters, sets it apart as a promising candidate for creating more balanced and fair machine learning models.

\section{Proposed Method}
\label{sec:method}
\begin{figure}[t!]
    \centering
    \includegraphics[width=0.88\textwidth]{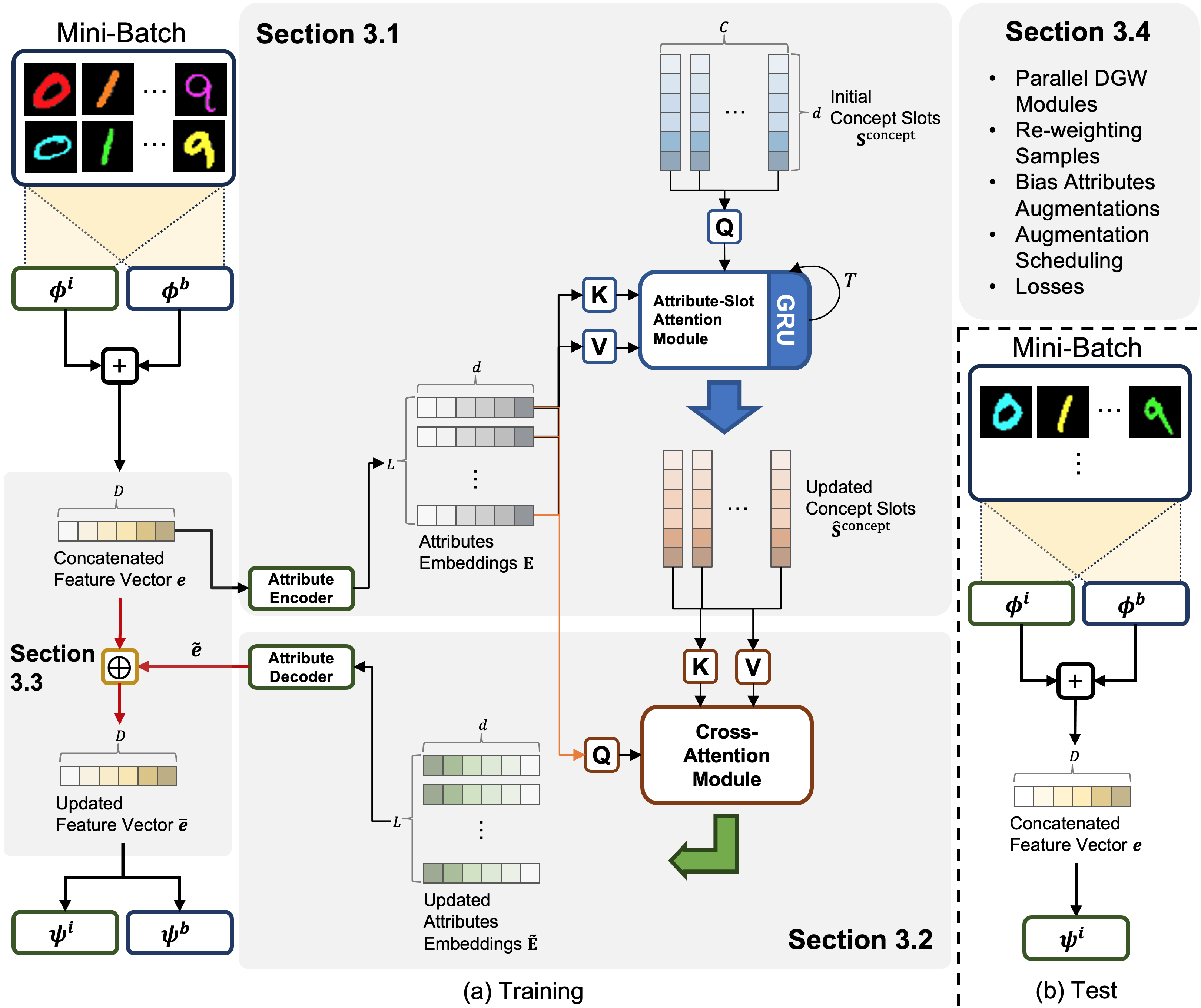}
    \caption{Overall architecture of Debiasing Global Workspace (DGW)}
    \label{fig:dgw_architecture}
    \vspace{-2em}
\end{figure}
Motivated by the observation in Section~\ref{subsec:ood_evaluation}, we propose a novel debiasing approach for attribution-centric representation learning, \emph{Debiasing Global Workspace} (DGW), that allows for learning the composition of attributes in a dataset and provides interpretable explanations about the model's decision-making, inspired by information bottleneck architecture~\cite{Hong_2024_WACV}. 

Since it has been shown in \cite{lee2021learning} the effectiveness of training the two separate encoders which embed an image into disentangled vectors corresponding to the intrinsic and bias attributes, respectively, the goal of our approach is to update the concatenated feature vector from two encoders to implicitly learn latent embeddings corresponding to intrinsic and bias attributes in a given dataset. Our model consists of: i) Attribute-Slot-Attention~(\texttt{ASA}) module~(Section~\ref{subsec:asa_module}) that extracts the latent concept embedding specific to each batch of input, ii) Cross-Attention~(\texttt{CA}) module~(Section~\ref{subsec:ca_module}) for broadcasting between input embedding and the extracted concept embedding from the \texttt{ASA} module so that it produces the updated attribute embedding, iii) Mixup strategy~(Section~\ref{subsec:mixup}) to mix and finalize the updated concatenated feature vector used to two classifiers, and iv) training schemes~(Section~\ref{subsec:training_schemes}). Overall architecture is described in Fig.~\ref{fig:dgw_architecture}, and additional details can be found in Appendix~\ref{app:method_detail}.

\subsection{Learning Latent Embedding of Attributes}
\label{subsec:asa_module}

From the intrinsic attribute encoder $\boldsymbol{\phi}^{i}$ and the bias attribute encoder $\boldsymbol{\phi}^{b}$, we derive a concatenated feature vector $\mathbf{e} = [ \boldsymbol{\phi}^{i}(\mathbf{x}); \boldsymbol{\phi}^{b}(\mathbf{x})] \in \mathbb{R}^{D}$ from input $\mathbf{x}$. The \texttt{Attribute Encoder} processes $\mathbf{e}$ into attribute embeddings $\mathbf{E} \in \mathbb{R}^{L \times d}$, acting as image patches or word tokens, using a simple linear model with layer normalization~\cite{ba2016layer} to lower computational complexity. The \texttt{Attribute-Slot-Attention (ASA)} module then applies competitive attention on $\mathbf{E}$, with concept slots $\mathbf{S}^{\mathtt{concept}} \in \mathbb{R}^{C \times d}$ generating an attention matrix $\mathbf{A}^{\mathtt{ASA}} \in \mathbb{R}^{C \times L}$ through linear projections $q^{\mathtt{ASA}}$, $k^{\mathtt{ASA}}$, and $v^{\mathtt{ASA}}$, and dot products, normalized to emphasize competition among concept slots.

We then seek to group and aggregate the attended attribute embedding and obtain the attention readout for each concept slot. Intuitively, this represents how much the attended attribute embedding contribute to semantically representing each concept. The process involves normalizing the attention matrix $\mathbf{A}^{\mathtt{ASA}}$ and multiplying it with $v^{\mathtt{ASA}}(\mathbf{E})$ to generate the attention readouts $\mathbf{U}$, which quantify each concept slot's semantic contribution. These readouts update concept slots through a Gated Recurrent Unit~(GRU)~\cite{cho2014learning}, refining the concept representation iteratively. The final concept slots are derived after several iterations, capturing the essence of each concept semantically. This streamlined approach facilitates efficient concept slot updating and representation refinement. This is described in Algorithm~\ref{alg:csa_module} in pseudo-code.

\subsection{Produce Updated Attribute Representations}\label{subsec:ca_module}
\begin{algorithm}[t!]
\caption{\textbf{Attribute-Slot-Attention~(ASA) module}. The module receives the attribute embedding $\mathbf{E} \in \mathbb{R}^{L \times d}$; the number of concepts $C$; and the dimension of concepts $d$. The model parameters include: the linear projection $q^{\mathtt{ASA}}, k^{\mathtt{ASA}}, v^{\mathtt{ASA}}$ with output dimension $d$; a \texttt{GRU} network; a Gaussian distribution's mean and diagonal covariance $\mu, \sigma \in \mathbb{R}^{d}$.}
\label{alg:csa_module}
\begin{algorithmic}
\State $\mathbf{S}^{\mathtt{concept}} = \mathtt{Tensor}(C, d)$
\Comment{$\mathtt{concept-slots} \in \mathbb{R}^{C \times d}$}
\State $\mathbf{S}^{\mathtt{concept}} \sim\mathcal{N}(\mu, \sigma)$
\State $\mathbf{E} = \mathtt{LayerNorm}(\mathbf{E})$
\For{$t = 0, \dots, T$}
    \State $\mathbf{S}^{\mathtt{concept}} = \mathtt{LayerNorm}(\mathbf{S}^{\mathtt{concept}})$
    \State $\mathbf{A}^{\mathtt{ASA}} = \operatorname{softmax}(\frac{1}{\sqrt{d}}q^{\mathtt{ASA}}(\mathbf{S}^{\mathtt{concept}}) \cdot k^{\mathtt{ASA}}(\mathbf{E})^{\top}, \mathtt{axis='concept-slots'})$
    \State $\mathbf{A}^{\mathtt{ASA}} = \mathbf{A}^{\mathtt{ASA}}\;/\;\mathbf{A}^{\mathtt{ASA}}\mathtt{.sum(axis='attribute-embedding')}$
    \State $\mathbf{U} = \mathbf{A}^{\mathtt{ASA}} \cdot v^{\mathtt{ASA}}(\mathbf{E})$
    \State $\mathbf{S}^{\mathtt{concept}} = \mathtt{GRU}(\mathtt{state} = \mathbf{S}^{\mathtt{concept}}, \mathtt{inputs} = \mathbf{U})$
\EndFor \\
\Return $\mathbf{S}^{\mathtt{concept}}$
\end{algorithmic}

\end{algorithm}
The Cross-Attention (\texttt{CA}) mechanism refines attribute embeddings $\mathbf{E} \in \mathbb{R}^{L \times d}$ by applying linear projections $q^{\mathtt{CA}}$, $k^{\mathtt{CA}}$, and $v^{\mathtt{CA}}$ to both $\mathbf{E}$ and concept slots $\hat{\mathbf{S}}$ with position embedding from the \texttt{ASA} module. This produces an attention matrix $\mathbf{A}^{\mathtt{CA}}$, indicating patch–concept slot relationships, and updates embeddings $\mathbf{\Tilde{\mathbf{E}}} = \mathbf{A}^{\mathtt{CA}} \cdot v^{\mathtt{CA}}(\hat{\mathbf{S}}^{\mathtt{concept}})$, showing attribute-concept correlations. $\mathbf{A}^{\mathtt{CA}}$ is key for interpretability as discussed in Section~\ref{subsec:interpretable_analysis}. The updated embeddings $\Tilde{\mathbf{E}}$ then input into the \texttt{Attribute Decoder}, producing a tentative feature vector $\boldsymbol{\Tilde{e}}$ of dimension $D$, ready to update the original feature vector $\mathbf{e}$.

\subsection{Mixup between Original and Updated Representations}
\label{subsec:mixup}
To produce the final attribute representation $\boldsymbol{\Bar{e}}$, there are various computational operations that can be applicable, such as the residual connection in ResNet~\cite{he2016deep}.
Instead, we introduce a modified version of Manifold Mixup\cite{verma2019manifold}. Manifold Mixup has demonstrated that interpolations in deeper hidden layers capture higher-level information effectively. While intermediate feature embeddings and one-hot labels from mini-batches are mixed in the original setup of Manifold Mixup, we perform to mix the feature embeddings only because the label information of those is the same in our definition. So, the final representation of embedding $\boldsymbol{\Bar{e}}$ is: $\mathbf{\Bar{e}} = \text{Mix}_{\alpha}(\mathbf{e}, \mathbf{\Tilde{e}})$
where $\text{Mix}_{\alpha} (a, b) = \alpha \cdot a + (1 - \alpha) \cdot b$, and the mixing coefficient $\alpha \sim \text{Beta}(\beta, \beta)$, as proposed in mixup~\cite{zhang2018mixup}. The updated feature vector $\mathbf{\Bar{e}}$ is fed to the classifiers $\boldsymbol{\psi}^{i}$ and $\boldsymbol{\psi}^{b}$.

\subsection{Training Strategies}
\label{subsec:training_schemes}
Our training scheme leverages two parallel modules, $\texttt{DGW}^{i}$ and $\texttt{DGW}^{b}$, for learning intrinsic and bias attributes, respectively. Each \texttt{DGW} module integrates \texttt{ASA} and \texttt{CA} mechanisms. Overall training strategies are detailed in Algorithm~\ref{alg:dgw_training_stragegies}.

\paragraph{Re-weighting Samples.}
We have two linear classifiers $\boldsymbol{\psi}^{i}$ and $\boldsymbol{\psi}^{b}$ that take the updated concatenated vector $\boldsymbol{\Bar{e}}$ from the \texttt{CA} module as input to predict the target label $y$. To train $\boldsymbol{\phi}^{i}, \texttt{DGW}^{i}$ and $\boldsymbol{\psi}^{i}$ as intrinsic feature extractor and $\boldsymbol{\phi}^{b}, \texttt{DGW}^{b}$ and $\boldsymbol{\psi}^{b}$ as bias extractor, we utilize the relative difficulty score of each data sample, proposed in~\cite{nam2020learning}.  Specifically, we train $\boldsymbol{\phi}^{b}, \texttt{DGW}^{b}$ and $\boldsymbol{\psi}^{b}$ to be overfitted to the bias attributes by utilizing the generalized cross entropy~(GCE)~\cite{zhang2018generalized}, while $\boldsymbol{\phi}^{i}, \texttt{DGW}^{i}$ and $\boldsymbol{\psi}^{i}$ are trained with the cross entropy~(CE) loss. Then, the samples with high CE loss from $\boldsymbol{\psi}^{b}$ can be regarded as the bias-conflicting samples compared to the samples with low CE loss. In this regard, we obtain the relative difficulty score of each data sample as:
\begin{equation}
    W(\boldsymbol{\Bar{e}}) = \frac{CE(\boldsymbol{\psi}^{b}(\boldsymbol{\Bar{e}}), y)}{CE(\boldsymbol{\psi}^{i}(\boldsymbol{\Bar{e}}), y) + CE(\boldsymbol{\psi}^{b}(\boldsymbol{\Bar{e}}), y)}
    \label{eq:resampling}
    \vspace{-0.5em}
\end{equation}

\paragraph{Bias Attributes Augmentation and Scheduling.}
\begin{algorithm}[t!]
\caption{\textbf{Training Strategies in Debiasing Global Workspace (DGW)}. image $\mathbf{x}$; label $y$; iteration $t$; and feature augmentation iteration $t_{\text{aug}}$ }
\label{alg:dgw_training_stragegies}
\begin{algorithmic}
\State Initialize two networks $(\boldsymbol{\phi}^{i}, \texttt{DGW}^{i}, \boldsymbol{\psi}^{i})$, $(\boldsymbol{\phi}^{b}, \texttt{DGW}^{b}, \boldsymbol{\psi}^{b})$
\While {not converged}
    \State Concatenate $\mathbf{e} = [\boldsymbol{\phi}^{i}(\mathbf{x});\boldsymbol{\phi}^{b}(\mathbf{x})]$
    \State Generate $\mathbf{\Bar{e}}$ by feeding $\mathbf{e}$ to $\texttt{DGW}^{i}, \texttt{DGW}^{b}$ and $\text{Mix}_{\alpha}$
    \State Update $(\boldsymbol{\phi}^{i}, \texttt{DGW}^{i}, \boldsymbol{\psi}^{i})$, $(\boldsymbol{\phi}^{b}, \texttt{DGW}^{b}, \boldsymbol{\psi}^{b})$ with $\mathcal{L}_{\text{re}} = W(\boldsymbol{\Bar{e}}) \cdot CE(\boldsymbol{\psi}^{i}(\boldsymbol{\Bar{e}}), y) + \lambda_{\text{re}}GCE(\boldsymbol{\psi}^{b}(\boldsymbol{\Bar{e}}), y) + \lambda_{\text{ent}}\mathcal{L}_{\text{ent}}$
    \If{$t > t_{\text{aug}}$}
        \State Randomly permute $\mathbf{e} = [\boldsymbol{\phi}^{i}(\mathbf{x});\boldsymbol{\phi}^{b}(\mathbf{x})]$ into $\mathbf{e}_{\text{swap}} = [ \boldsymbol{\phi}^{i}(\mathbf{x});  \boldsymbol{\phi}^{b}_{\text{swap}}(\mathbf{x})]$
        \State Generate $\mathbf{\Bar{e}}_{\text{swap}}$ by feeding $\mathbf{e}_{\text{swap}}$ to $\texttt{DGW}^{i}, \texttt{DGW}^{b}$ and $\text{Mix}_{\alpha}$
        \State Update $(\boldsymbol{\phi}^{i}, \texttt{DGW}^{i}, \boldsymbol{\psi}^{i})$, $(\boldsymbol{\phi}^{b}, \texttt{DGW}^{b}, \boldsymbol{\psi}^{b})$ with $\mathcal{L}_{\text{total}} = \mathcal{L}_{\text{re}} + \lambda_{\text{swap}}\mathcal{L}_{\text{swap}}$
    \EndIf
\EndWhile
\end{algorithmic}
\end{algorithm}
\vspace{-0.5em}
Following~\cite{lee2021learning}, we also leverage the augmentation method of bias-conflicting samples by swapping the disentangled latent vectors among the training sets. We randomly permute the intrinsic and bias features in each mini-batch and obtain $\mathbf{e}_{\text{swap}} = [ \boldsymbol{\phi}^{i}(\mathbf{x});  \boldsymbol{\phi}^{b}_{\text{swap}}(\mathbf{x})]$ where $\boldsymbol{\phi}^{b}_{\text{swap}}(\mathbf{x})$ denotes the randomly permuted bias attributes of $\boldsymbol{\phi}^{b}(\mathbf{x})$. $\mathbf{e}_{\text{swap}}$ acts as augmented bias-conﬂicting latent vectors with diversity inherited from the bias-aligned samples. Since the goal of our method is to update the concatenate feature vector, $\mathbf{e}_{\text{swap}}$ is also fed to \texttt{DGW} module, and the updated representation $\mathbf{\Tilde{e}_{\text{swap}}}$ is produced, as similar to $\mathbf{\Tilde{e}}$. Thus, $\mathbf{\Bar{e}}_{\text{swap}} = \text{Mix}_{\alpha}(\mathbf{e}_{\text{swap}}, \mathbf{\Tilde{e}}_{\text{swap}})$. Furthermore, We leverage a training schedule of feature augmentation, ensuring that the augmentation should only be applied after they are disentangled to a certain degree.

\paragraph{Final Loss.}
Referring to~\cite{Hong_2024_WACV}, we use the entropy loss $\mathcal{L}_{\text{ent}} = H(\mathbf{A}^{\mathtt{CA}}_{i}) + H(\mathbf{A}^{\texttt{CA}}_{b})$ where $H(\mathbf{A}) = H(a_{1}, \cdots , a_{|\mathbf{A}|}) = (1/|\mathbf{A}|)\sum_{i} -a_{i} \cdot \log(a_{i})$ to minimize the entropy of the attention masks from the \texttt{CA} modules. Thus, the objective function can be written as:
\begin{equation}
    \mathcal{L}_{\text{re}} = W(\boldsymbol{\Bar{e}}) \cdot CE(\boldsymbol{\psi}^{i}(\boldsymbol{\Bar{e}}), y) + \lambda_{\text{re}}GCE(\boldsymbol{\psi}^{b}(\boldsymbol{\Bar{e}}), y) + \lambda_{\text{ent}}\mathcal{L}_{\text{ent}}
    \label{eq:re_loss}
\end{equation}

To ensure that $\boldsymbol{\psi}^{i}$ and $\boldsymbol{\psi}^{b}$ predicts target labels mainly based on $\boldsymbol{\Bar{e}}$, respectively, the loss from $\boldsymbol{\psi}^{i}$ is not backpropagated to $\boldsymbol{\phi}^{b}$ and $\texttt{DGW}^{b}$, and vice versa.
Along with $\mathcal{L}_{\text{re}}$, we add the following loss function to train two neural networks with the augmented features: 
\begin{equation}
   \mathcal{L}_{\text{swap}} = W(\boldsymbol{\Bar{e}}) \cdot CE(\boldsymbol{\psi}^{i}(\boldsymbol{\Bar{e}}_{\text{swap}}), y) + \lambda_{\text{re}}GCE(\boldsymbol{\psi}^{b}(\boldsymbol{\Bar{e}}_{\text{swap}}), \Tilde{y})
   \label{eq:swap_loss}
\end{equation}
where $\Tilde{y}$ denotes target labels for permute bias attributes $\boldsymbol{\phi}^{b}_{\text{swap}}(\mathbf{x})$. Therefore, our total loss function is described as: $\mathcal{L}_{\text{total}} = \mathcal{L}_{\text{re}} + \lambda_{\text{swap}} \cdot \mathcal{L}_{\text{swap}}$ where $\lambda_{\text{swap}}$ is adjusted for weighting the importance of the feature augmentation.

\section{Experiments}
\label{sec:exp}
In this section, we describe experimental results about i) performance evaluation on various biased datasets (Section~\ref{subsec:performance_evaluation}), ii) interpretable analysis for attribute-centric representation learning (Section~\ref{subsec:interpretable_analysis}), and iii) additional qualitative and quantitative analyses (Section~\ref{subsec:additional_analysis}).

Following~\cite{lee2021learning}, we use two synthetic datasets, Colored MNIST~(C-MNIST) and Corrputed CIFAR10~(C-CIFAR-10), and one real-world dataset Biased FFHQ (BFFHQ), to evaluate the generalization of debiasing baselines over various domains in Section~\ref{subsec:performance_evaluation} and Section~\ref{subsec:interpretable_analysis}. Curated from the FFHQ dataset~\cite{karras2019style}, BFFHQ includes human face images annotated with their facial attributes. We selected age and gender as the intrinsic and bias attributes among the facial attributes, respectively, and formed the dataset with images of high correlation between them. More specifically, most of the females are `young' (i.e., ages 10 to 29), and males are `old' (i.e., ages 40 to 59). Thus, bias-aligned samples composing most datasets are young women and old men.

We adjusted the degree of correlation for each dataset by varying the number of bias-conflicting samples in the training dataset. Specifically, we set the ratio of bias-conflicting samples to 0.5\%, 1\%, 2\%, and 5\% for both C-MNIST and C-CIFAR-10, respectively, and to 0.5\% for BFFHQ. We then constructed an unbiased test set for C-MNIST and C-CIFAR-10, which included images that did not exhibit a high correlation in the training set. For BFFHQ, we constructed a bias-conflicting test set, which excluded the bias-aligned samples from the unbiased test set. This was necessary because the bias-aligned images comprised half of the unbiased test set in BFFHQ and could still be correctly classified by the biased classifier. This could result in an unintentional inflation of the accuracy of the unbiased test set. Therefore, we intentionally used the bias-conflicting test set for BFFHQ.

\subsection{Performance Evaluation}
\label{subsec:performance_evaluation}

\paragraph{Baselines.}
Our set of baselines includes six different approaches~\footnote{We only establish baselines that can be directly tested. For example, $\chi^{2}$~\cite{zhang2023learning} is also relevant to our work, but its code is not publicly available.}: Vanilla network, HEX~\cite{wang2018learning}, EnD~\cite{tartaglione2021end}, ReBias~\cite{bahng2020learning}, LfF~\cite{nam2020learning}, and LFA~\cite{lee2021learning}. Vanilla refers to the classification model trained only with the original cross-entropy~(CE) loss without debiasing strategies. EnD leverages the explicit bias labels, such as the color labels in C-MNIST dataset, during the training phase. HEX and ReBias assume an image's texture as a bias type, whereas LfF, LFA, and our method do not require any prior knowledge about the bias type.

\paragraph{Implementation Details.}
We mainly follow the implementation details from~\cite{lee2021learning}. As an encoder, we use a fully connected network with three hidden layers for C-MNIST and ResNet-18~\cite{he2016deep} for the remaining datasets. Since the disentangled vectors from the intrinsic and bias encoders must be handled together, we leverage a fully connected classifier by doubling the number of hidden units.
As shown in Fig.~\ref{fig:dgw_architecture}(b), during the test phase, we use the intrinsic classifier $\boldsymbol{\psi}^{i}(\mathbf{e})$ for the ﬁnal prediction only, where $\mathbf{e} = [ \boldsymbol{\phi}^{i}(\mathbf{x});  \boldsymbol{\phi}^{b}(\mathbf{x})]$. We utilized a batch size of 256 for C-MNIST and C-CIFAR-10, and of 64 for BFFHQ during the training process. The number of concepts $C$ and the size of $d$ in the \texttt{ASA} module are set to 2 and 8 for C-MNIST. For C-CIFAR-10, $C$ and $d$ are set to 5 and 16, whereas 10 and 32 for BFFHQ. We applied bias attribute augmentation after 10K iterations for all datasets. We conducted experimental trials with three different random seeds and reported the average performance of the unbiased test sets with the mean and standard deviation. More implementation details are included in Appendix~\ref{subapp:performance}.
\begin{table}[t!]
\centering
\caption{Test accuracy evaluated on unbiased test sets of C-MNIST and C-CIFAR-10, and the bias-conﬂicting test set of BFFHQ with varying ratio of bias-conﬂicting samples. For HEX and EnD, performance from~\cite{lee2021learning}. For Vanilla, ReBias, LfF, LFA and DGW, results from our evaluation. \cmark/\xmark \;indicates debiasing methods with/without predefined bias types, respectively.}
\scalebox{0.88}{
\begin{tabular}{c|c|m{4em}|m{4em}|m{4em}|m{4em}|m{4em}|m{4em}|m{5.5em}}
 \toprule
 \multirow{2}{3em}{Dataset} & \multirow{2}{4em}{Ratio~(\%)} & Vanilla & HEX & EnD & ReBias & LfF & LFA & DGW~(Ours) \\
 \cline{3-9}
 & & \xmark & \cmark & \cmark & \cmark & \xmark & \xmark & \xmark \\
 \hline
 \multirow{4}{*}{C-MNIST} & 0.5 & $36.2_{\pm 1.8}$ & $30.3_{\pm 0.8}$ & $34.3_{\pm 1.2}$ & $\textbf{72.2}_{\pm 1.5}$ & $47.5_{\pm 3.0}$ & $67.4_{\pm 1.7}$ & $68.9_{\pm 2.8}$ \\
 & 1.0 & $50.8_{\pm 2.3}$ & $43.7_{\pm 5.5}$ & $49.5_{\pm 2.5}$ & $\textbf{86.6}_{\pm 0.6}$ & $64.6_{\pm 2.5}$ & $79.0_{\pm 1.0}$ & $81.3_{\pm 1.2}$ \\
 & 2.0 & $65.2_{\pm 2.1}$ & $56.9_{\pm 2.6}$ & $68.5_{\pm 2.2}$ & $\textbf{92.7}_{\pm 0.3}$ & $74.9_{\pm 3.7}$ & $85.0_{\pm 0.8}$ & $84.6_{\pm 1.5}$ \\
 & 5.0 & $81.6_{\pm 0.6}$ & $74.6_{\pm 3.2}$ & $81.2_{\pm 1.4}$ & $\textbf{97.1}_{\pm 0.6}$ & $80.2_{\pm 0.9}$ & $88.7_{\pm 1.3}$ & $88.9_{\pm 0.2}$ \\
 \hline
 \multirow{4}{*}{C-CIFAR-10} & 0.5 & $22.8_{\pm 0.3}$ & $13.9_{\pm 0.1}$ & $22.9_{\pm 0.3}$ & $20.8_{\pm 0.2}$ & $25.0_{\pm 1.5}$ & $27.9_{\pm 1.0}$ & $\textbf{29.0}_{\pm 1.9}$ \\
 & 1.0 & $26.2_{\pm 0.5}$ & $14.8_{\pm 0.4}$ & $25.5_{\pm 0.4}$ & $24.4_{\pm 0.4}$ & $31.0_{\pm 0.4}$ & $34.3_{\pm 0.6}$ & $\textbf{34.9}_{\pm 0.4}$ \\
 & 2.0 & $31.1_{\pm 0.6}$ & $15.2_{\pm 0.5}$ & $31.3_{\pm 0.4}$ & $29.6_{\pm 2.9}$ & $38.3_{\pm 0.4}$ & $40.3_{\pm 2.4}$ & $\textbf{41.1}_{\pm 0.9}$ \\
 & 5.0 & $42.0_{\pm 0.3}$ & $16.0_{\pm 0.6}$ & $40.3_{\pm 0.9}$ & $41.1_{\pm 0.2}$ & $48.8_{\pm 0.9}$ & $50.3_{\pm 1.1}$ & $\textbf{51.3}_{\pm 0.7}$ \\
 \hline
 BFFHQ & 0.5 & $54.5_{\pm 0.6}$ & $52.8_{\pm 0.9}$ & $56.9_{\pm 1.4}$ & $58.0_{\pm 0.2}$ & $63.6_{\pm 2.9}$ & $59.5_{\pm 3.8}$ & $\textbf{63.7}_{\pm 0.6}$ \\
 \bottomrule
\end{tabular}}
\label{tab:test_performance}
\end{table}

\paragraph{Performance Comparison.}
Table~\ref{tab:test_performance} provides a comprehensive comparison of test performance among baselines. ReBias stands out on C-MNIST dataset, outperforming our DGW method. This can be attributed to its specialized design. Despite its effectiveness, ReBias's necessity for predefined bias categories constrains its versatility, especially when dealing with datasets where bias types are not readily identifiable. In contrast, DGW method does not require any foreknowledge of the biases present. Furthermore, when compared to LFA, which serves as our direct benchmark, DGW approach demonstrates improved performance across all datasets. This underscores the effectiveness of DGW, without the need for predefined bias specifications, making it a robust and flexible architecture for debiasing in image classification tasks.

\subsection{Analysis for Attribute-Centric Representation Learning}
\label{subsec:interpretable_analysis}
This section presents an analysis of shape-centric representation learning, which allows for visualizing the behavior of intrinsic and biased image attributes via the \texttt{CA} module of Section~\ref{subsec:ca_module} in our approach.
Since we trained two \texttt{DGW} modules, the model generates two attention masks: $\mathbf{A}^{\mathtt{CA}}_{i}$ for intrinsic attributes, focusing on essential object features like shape or structure, and $\mathbf{A}^{\mathtt{CA}}_{b}$ for biased attributes, which capture non-essential features like color or texture. 
Visualizing the masks aids in identifying and mitigating potential biases, ensuring classification decisions are based on relevant features.

\begin{figure*}[t!]
	\centering
	\includegraphics[width=0.8\linewidth]{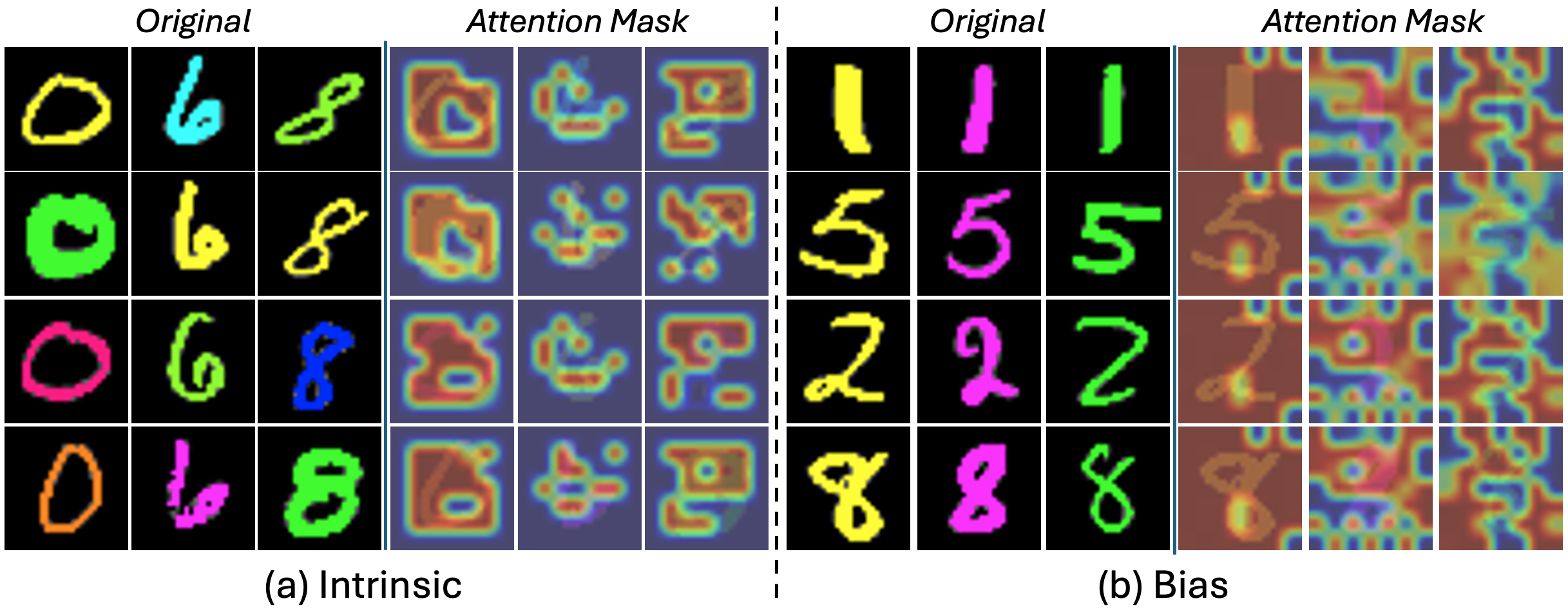}
 \vspace{-0.5em}
        \caption{Visualization of attention masks generated by intrinsic and bias components of DGW for the C-MNIST dataset}
        \label{fig:cmnist_mask}	
\vspace{-1em}
\end{figure*}

For visualization purposes, the number of concepts $C$ in the \texttt{ASA} module are set to 2 for C-MNIST and set to 10 for C-CIFAR-10. Our experiments with C-MNIST and C-CIFAR-10 datasets reveal the model's adeptness in distinguishing intrinsic from biased attributes. In C-MNIST, the intrinsic attention masks $\mathbf{A}^{\mathtt{CA}}_{i}$ spotlighted digit shapes, deliberately overlooking color variations. This is particularly evident in digits such as 0, 6, and 8, where the intrinsic model's attention on shape over color showcases its ability to overcome bias-inducing attributes, as illustrated in Fig. \ref{fig:cmnist_mask}(a). Conversely, the biased attention masks $\mathbf{A}^{\mathtt{CA}}_{b}$ display a strong emphasis on color. This is evident in Fig. \ref{fig:cmnist_mask}(b), where the attention mask for digits of the same color appears nearly identical. This clear distinction underlines DGW's refined approach to identifying and managing biased versus intrinsic attributes.

\begin{figure*}[t!]
	\centering
	\includegraphics[width=0.65\linewidth]{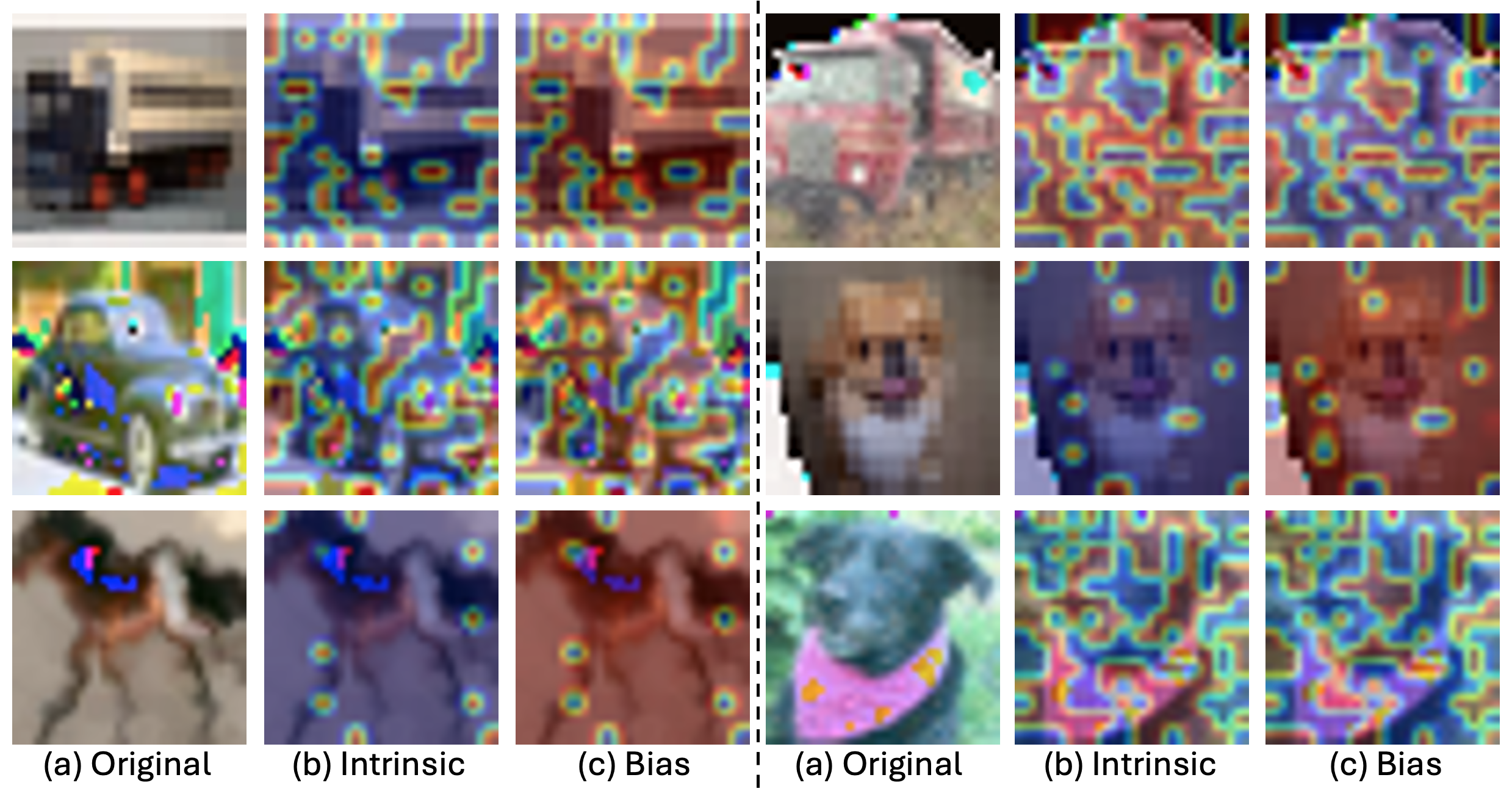}
 \vspace{-0.5em}
        \caption{Visualization of attention masks generated by intrinsic and bias components of DGW for the C-CIFAR10 dataset}
        \label{fig:cifar_mask}
\vspace{-2em}
\end{figure*}

In our analysis of the C-CIFAR-10 dataset, a similar pattern emerged. The intrinsic attention masks $\mathbf{A}^{\mathtt{CA}}_{i}$ were observed to focus selectively on parts of the image that were uncorrupted, effectively ignoring any corrupted sections. This selectiveness highlights the model's ability to concentrate on relevant image features for classification. In contrast, the biased attention masks $\mathbf{A}^{\mathtt{CA}}_{b}$ did not make distinctions between corrupted and uncorrupted parts of the image, instead covering all image attributes without discrimination. This behavior illustrates the near-perfect inverse correlation between the intrinsic and biased components' attention masks, further underscoring the model's proficiency in distinguishing between essential and non-essential image attributes for informed decision-making. The dynamic interplay between two attention masks is captured in Fig. \ref{fig:cifar_mask}, particularly in parts (b) and (c), where their complementary roles are demonstrated, highlighting the model's nuanced approach.
This granular understanding and clear distinction between intrinsic and biased attention underscore the model's refined skill in navigating and utilizing the various features presented in images. It illustrates a deliberate allocation of focus that not only enhances classification accuracy but also actively reduces the model's reliance on biased or non-essential features. For more detailed analysis and additional experimental findings, refer to Appendix~\ref{subapp:attribute_centric}

\subsection{Quantitative and Qualitative Analysis}
\label{subsec:additional_analysis}
In this section, we present additional quantitative and qualitative analysis to compare our method to Vanilla, and LFA~\cite{lee2021learning}. More experimental results with different settings can be found in Appendix~\ref{subapp:additional_analysis}.

\begin{figure*}[t!]
	\centering
	\includegraphics[width=0.95\linewidth]{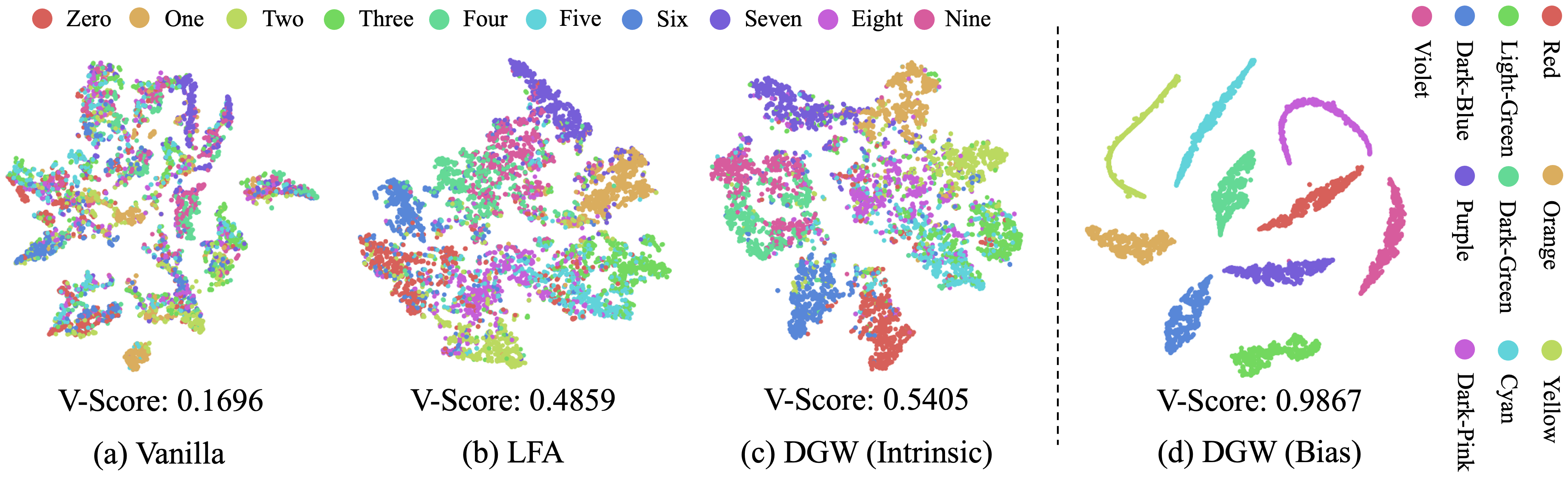}
  \vspace{-0.5em}
	\caption{t-SNE plots for intrinsic and bias features on C-MNIST (with 0.5$\%$ setting). }\label{fig:tsne_plot}
\vspace{-0.3em}
\end{figure*}

\paragraph{t-SNE and Clustering.}
We measure the clustering performance with t-SNE \cite{vandermaaten08a} and V-Score \cite{rosenberg2007v} of features from various models capturing intrinsic and bias attributes on C-MNIST, where V-Score represents homogeneity and completeness, implying a higher value is better clustering. In Fig. \ref{fig:tsne_plot}, the intrinsic encoder $\boldsymbol{\phi}^{i}$ in our method for capturing intrinsic attributes encourages to learn the disentangled features to get tighter clusters and better separation, implying better performance for intra- and inter-classification tasks, compared to baselines, which is presented with V-Score.
Also, bias attributes are well captured by the bias encoder $\boldsymbol{\phi}^{b}$ in our method as shown in Fig. \ref{fig:tsne_plot}(d). 
\begin{figure*}[t!]
	\centering
	\includegraphics[width=0.99\linewidth]{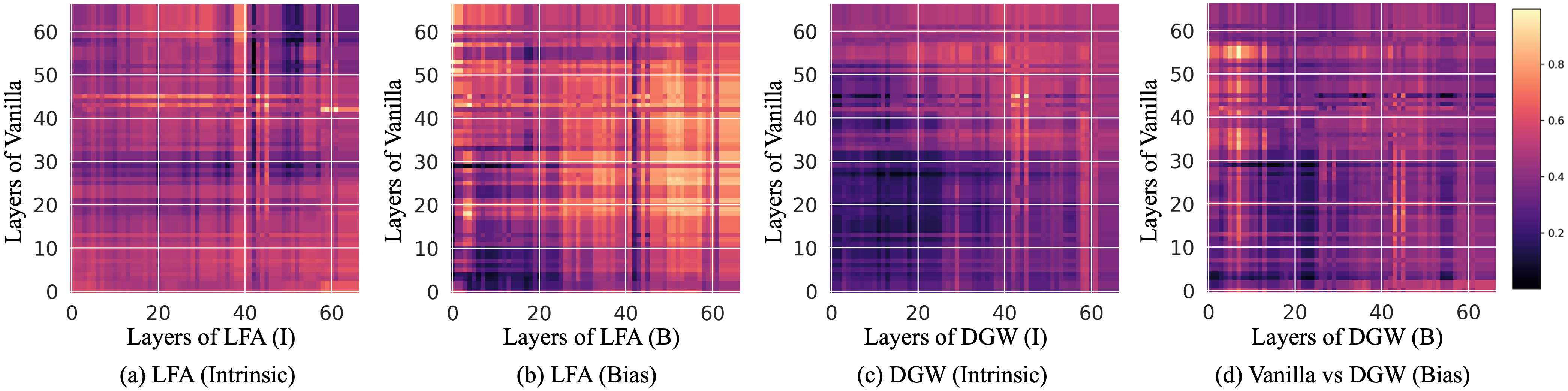}
 \vspace{-0.5em}
	\caption{Representations of similarities for vanilla and different methods with all pairs of layers on C-CIFAR-10 (0.5$\%$ setting). High similarity score denotes high values. }\label{fig:model_sim_cifar05}
 \vspace{-1em}
\end{figure*}

\paragraph{Model Similarity.}
To analyze the behavior of models, we visualize model similarity representations with Centered Kernel Alignment (CKA)\cite{raghu2021vision, kornblith2019similarity, cortes2012algorithms}, showing similarities between all pairs of layers with different models. Note, I and B denote intrinsic and bias attribute encoders, representation map is normalized to present similarity or difference more intuitively, and $\boldsymbol{\phi}^{b}$ and $\boldsymbol{\phi}^{i}$ in our approach are used for comparison. As illustrated in Fig. \ref{fig:model_sim_cifar05}, Vanilla and LFA possess similar weights in many layers regardless of intrinsic and bias attributes, represented with bright colors.
The deeper layers show more similarities than the initial layers.
On the other hand, Vanilla and DGW show prominently less similarity values for the initial as well as the deeper layers, compared to the baselines across different datasets and settings.
This implies that DGW and baselines have different behaviors across layers of a model. 

\paragraph{Model Reliability.}
\begin{wraptable}{R}{8cm} 
\vspace{-1em}
\caption{ECE ($\%$) and NLL under different settings on C-CIFAR-10.}
\vspace{-0.5em}
\begin{center}
\scalebox{0.63}{\begin{tabular}{l|cc|cc|cc|cc}
\toprule
Ratio ($\%$) & \multicolumn{2}{c}{0.5} & \multicolumn{2}{c}{1.0} & \multicolumn{2}{c}{2.0} & \multicolumn{2}{c}{5.0} \\ 
& ECE &  NLL & ECE & NLL & ECE & NLL & ECE & NLL  \\ 
\hline
Vanilla & 13.75 & 5.99 & 13.14 & 9.87 & 12.25 & 6.65 & 13.76 & 5.99 \\
LFA & 12.09 & 5.81& \textbf{11.45} & 7.27 & 10.25 & 5.14 & 7.56 & 3.09 \\
DGW (Ours) & \textbf{11.85} & \textbf{5.71} & 11.53 & \textbf{6.88} & \textbf{9.96} & \textbf{4.41} & \textbf{7.55} & \textbf{3.01} \\
\bottomrule
\end{tabular}}
\label{tab:reliability_table}
\end{center}
\vskip -0.1in
\end{wraptable}
We computed Expected Calibration Error (ECE) and Negative Log Likelihood~(NLL) \cite{guo2017calibration} to explore model generalizability, where ECE is to measure calibration error and NLL is to calculate probabilistic quality of a model.
As explained in Table \ref{tab:reliability_table}, in overall cases, DGW shows the lowest ECE and NLL, implying a better model generalizability compared to baselines.

\section{Related Work}
\label{sec:related_work}
To reduce bias in machine learning, a tailored network can be designed for specific types of bias~\cite{tu2022multiattribute, zhao2020attribute, agarwal2020towards, bahng2020learning, geirhos2018imagenet, goel2020model, kim2019learning, li2020shape, minderer2020automatic, tartaglione2021end, wang2020towards}. However, this requires manual annotations and grouping of attribute types, which can be time-consuming. 
Based on the observation that malicious bias attributes are more accessible to learn than intrinsic ones~\cite{nam2020learning}, many approaches have been proposed using dynamic training schemes, such as re-weighing samples, and augmentation methods\cite{geirhos2018imagenet, lee2021learning, minderer2020automatic, li2019repair, lim2023biasadv}. However, the limited number of such samples does not help in learning meaningful representations, limiting  becomes a strong constraint on debiasing learning. Some studies with invariant feature learning~\cite{akuzawa2020adversarial, ganin2016domain} and representation rectification have been proposed for debiasing~\cite{zhang2023learning}, but they require complex model designs and complicated training schemes. 

Current methods need to understand how models perform under different attributes and require more analysis to distinguish intrinsic from biased properties. Since these distinctions serve as crucial cues for enhancing performance in debiasing, they are required to be addressed with great importance and further explored~\cite{dong2017improving, chen2019looks, chen2020concept, haar2023analysis}. Our method can explain how the model effectively pays attention to such attributes and learning representations through an attention-based module without bias annotation.

\section{Conclusion}
\label{sec:conclusion}
In this work, we introduced Debiasing Global Workspace, robust framework designed to learn debiased representations of attributes. Our method ensures improved performance and offers interpretable explanations by utilizing attention masks to capture intrinsic and biased attributes. Our approach yields tighter clusters and better model separation compared to baselines. Also, our method demonstrated superior model reliability and generalizability. For future research direction, we envision extending our framework into a general-purpose drop-in layer, aiming to enhance robust performance across recognition tasks.

\bibliographystyle{apalike}
\bibliography{preprint_arXiv/preprint_arXiv}  

\clearpage

\appendix
\section*{Appendix}
\label{sec:appendix}
\renewcommand\thefigure{A-\arabic{figure}}\setcounter{figure}{0}
\renewcommand\thetable{A-\arabic{table}}
\setcounter{table}{0}
\setcounter{section}{0}
\pagenumbering{Alph}

\section{Reproducibility}
\label{app:reprod}
All source codes, figures, models, etc., are available at \url{https://github.com/jyhong0304/debiasing_global_workspace}.

\section{Analysis Details}
\label{app:analysis}
In this section, we provide details of our observatory analysis to verify the shape-centric representation learning and robust performance on out-of-distribution (OOD) datasets of the Concept-Centric Transformer (CCT)~\cite{Hong_2024_WACV}. 

\subsection{Implementation Details}
\label{subapp:implementation_detail}

\paragraph{In Section~\ref{subsec:xai_cct}.}
We tried to validate CCT's shape-centric representation and latent concept learnings following the experimental setup on the ImageNet dataset with 200 classes in~\cite{Hong_2024_WACV}. We chose the combination of small-sized Vision Transformer (ViT-S)~\cite{dosovitskiy2020image} as
a backbone architecture and BO-QSA~\cite{jia2022improving} as a memory module because that pair achieved the best performance shown in~\cite{Hong_2024_WACV}.  
The hyperparameters we set are as follows:
\begin{itemize}
    \item Batch size: 256
    \item Training epochs: 10
    \item Warmup Iterations: 10
    \item Explanation loss $\lambda$: 0.
    \item  Weight decay: 1e-3
    \item Attention sparsity: 0.
    \item Number of concepts $C$: 50
\end{itemize}

\paragraph{In Section~\ref{subsec:ood_evaluation}.}
We trained CCT on ImageNet-1K following the experimental setup in the supplementary material in~\cite{Hong_2024_WACV}. We configured the ViT-S as the backbone and BO-QSA as the memory module again. The hyperparameter settings are the same as above, except that the number of $C$ concepts is 150.

\subsection{Experimental Design of OOD Benchmarking}
In~\cite{geirhos2021partial}, a Python toolbox was proposed to assess human and machine vision gaps. The toolbox enables evaluating PyTorch\texttrademark and TensorFlow\texttrademark models on 17 out-of-distribution datasets with high-quality human comparison data. 

For variants of ResNet~\cite{he2016deep} and WideResNet (WRN) \cite{zagoruyko2016wide}, we leverage \href{https://pytorch.org/docs/1.4.0/torchvision/models.html}{PyTorch Torchvision Model Zoo} that the toolbox supports to load the pretrained ResNet models. The list of the ResNet variants is as follows: ResNet-18, ResNet-34, ResNet-50, ResNet-101, WRN-50-2, WRN-101-2.

\section{Method Details}
\label{app:method_detail}
In this section, we describe the details of our proposed method, Debiasing Global Workspace (DGW).

\subsection{Background}
\label{subapp:background}

\paragraph{Global Workspace Theory.}
In neuroscience and cognitive science, there is an ongoing effort to develop theories of consciousness (ToCs) to identify the neural correlates of consciousness (Check the various ToCs in~\cite{seth2022theories}). One such theory is the Global Workspace Theory (GWT)~\cite{baars1993cognitive,dehaene2011experimental,mashour2020conscious}, which draws inspiration from the `blackboard' architecture used in artificial intelligence. In this architecture, a centralized resource, the blackboard, facilitates information sharing among specialized processors. 

In artificial intelligence, it is commonly believed that creating an intelligent system from several interacting specialized modules is more effective than building a single `monolithic' entity that can handle a wide range of conditions and tasks~\cite{goyal2022inductive, minsky1988society, robbins2017modularity}. Therefore, there has been a significant effort to synchronize these specialized modules with a centralized resource through a Shared Global Workspace (SGW)~\cite{goyal2021coordination} inspired by GWTs. The SGW is one such theory that shows how the attention mechanism can encourage the sharing of the most helpful information among neural modules in modern AI frameworks and is applied to various domains~\cite{dehaene2011experimental, goyal2019recurrent, jaegle2021perceiver, munkhdalai2019metalearned, santoro2018relational, Hong_2024_WACV}. Especially, Concept-Centric Transformer (CCT)~\cite{Hong_2024_WACV} applies the SGW to build explainable AI by utilizing an explicit working memory based on object-centric representation learning to improve the generalization of transformer-based models in the context of explainable models.

Thus, the name of our method, Debiasing Global Workspace, originates from the above theories. Our method applies the GWT/SGW to build a debiasing method. It allows a model to have a modular architecture with a working memory module (the \texttt{ASA} module in Section~\ref{subsec:asa_module} in the main text) and to update better attribute embedding and representation vectors (the \texttt{CA} module in Section~\ref{subsec:ca_module} in the main text). 

\paragraph{Object-Centric Representation Learning.}
Humans still outperform the most sophisticated AI technologies due to our exceptional ability to recombine previously acquired knowledge, enabling us to extrapolate to novel scenarios~\cite{fodor1988connectionism, goyal2022inductive, greff2020binding}.
Pursuing learning representations that can be effectively generalized compositionally has been a significant area of research, and one prominent effort that has emerged in this area is object-centric representation learning~\cite{burgess2019monet, greff2019multi, locatello2020object, chang2022object, jia2022improving}. This approach is designed to represent each object in an image using a unique subset of the image's latent code. Due to their modular structure, object-centric representations can effectively enable compositional generalization.

Due to its simple yet effective design, Slot-Attention (SA)~\cite{locatello2020object} has gained significant attention in unsupervised object-centric representation learning. The iterative attention mechanism allows SA to learn and compete between slots for explaining parts of the input, showing a soft clustering effect on visual inputs~\cite{locatello2020object}. Recently, many variants of slot-based methods have been proposed to overcome some inherent limitations, such as I-SA~\cite{chang2022object} and BO-QSA~\cite{jia2022improving}. Based on those methods, Concept-Centric Transformer (CCT)~\cite{Hong_2024_WACV} is one of several existing examples of leveraging slot-based methods in explainable models to extract semantic concepts. 

In the field of debiasing methods, few studies have emphasized the perspective of learning decomposable latent representations for intrinsic and biased attributes. Our approach emphasizes the potential of compositional generalization in debiasing learning, utilizing the slot-based method to implement a crucial module (Section~\ref{subsec:asa_module} in the main text). The benefits of this method are noteworthy and deserve further exploration.

\subsection{Algorithm Details}
Algorithm~\ref{alg:csa_module} in the main text shows the details of the \texttt{ASA} module in our method in pseudo-code. Algorithm~\ref{alg:csa_module} is described based on the SA~\cite{locatello2020object}, but we simplify it by removing the last \texttt{LayerNorm} and \texttt{MLP} layers. 

In~\cite{Hong_2024_WACV}, some variants of the slot-based method were leveraged, including SA, I-SA~\cite{chang2022object}, and BO-QSA~\cite{jia2022improving}, as a working memory module. However, we also used BO-QSA only because it contributed to the best performance across all experimental settings in ~\cite{Hong_2024_WACV}. We omit details of their technical differences as they are out of scope.



\subsection{Number of Iterations in the \texttt{ASA} module}
In the original usage of slot-based approaches, the refinement could be repeated several times depending on the tasks~\cite{chang2022object, locatello2020object}.
For setting the number of iterations $T$~(Algorithm~\ref{alg:csa_module} in the main text), we searched for the optimal number of iterations $T$ for each task as one of the hyperparameters. The setup is described in Section~\ref{subapp:performance}.



\section{Further Experimental Results and Details}
\label{app:exp}
In this section, we explain further experimental results and details. All experiments are conducted with three different random seeds and $95\%$ confidence intervals. 

\subsection{Hardware Specification of The Server}
The hardware specification of the server that we used to experiment is as follows:
\begin{itemize}
    \item CPU: Intel\textregistered{} Core\textsuperscript{TM} i7-6950X CPU @ 3.00GHz (up to 3.50 GHz)
    \item RAM: 128 GB (DDR4 2400MHz)
    \item GPU: NVIDIA GeForce Titan Xp GP102 (Pascal architecture, 3840 CUDA Cores @ 1.6 GHz, 384-bit bus width, 12 GB GDDR G5X memory)
\end{itemize}

\subsection{Datasets}
\label{subapp:datasets}
We describe the details of biased datasets, Colored MNIST (C-MNIST), Corrupted CIFAR-10 (C-CIFAR-10), and BFFHQ.

\paragraph{Colored MNIST.}
As per the existing studies~\cite{nam2020learning, kim2019learning, li2019repair, bahng2020learning, darlow2020latent, lee2021learning}, this biased dataset comprises two attributes that are highly correlated with each other-\emph{color} and \emph{digit}. To create this dataset, we added specific colors to the foreground of each digit, following~\cite{nam2020learning, darlow2020latent}. We generate a total of bias-aligned and bias-conflicting samples for different ratios of bias-conflicting samples:
\begin{itemize}
    \item $0.5\%$: (54751:249)
    \item $1\%$: (54509:491)
    \item $2\%$: (54014:986)
    \item $5\%$: (52551:2449)
\end{itemize}

\paragraph{Corrupted CIFAR-10.}
Among 15 different corruptions introduced in the original dataset~\cite{hendrycks2018benchmarking}, we selected corruption types for the Corrupted CIFAR-10 dataset, including \emph{Brightness, Contrast, Gaussian Noise, Frost, Elastic Transform, Gaussian Blur, Defocus Blur, Impulse Noise, Saturate}, and \emph{Pixelate}. These types of corruption are closely related to the original classes of CIFAR-10~\cite{krizhevsky2009learning}: \emph{Plane, Car, Bird, Cat, Deer, Dog, Frog, Horse, Ship}, and \emph{Truck}. For the dataset, we used the most severe level of corruption from the five different levels of corruption described in~\cite{hendrycks2018benchmarking}. We provide the total number of images of bias-aligned and bias-conflicting samples for each ratio of bias-conflicting samples:
\begin{itemize}
    \item $0.5\%$: (44832:228)
    \item $1\%$: (44527:442)
    \item $2\%$: (44145:887)
    \item $5\%$: (42820:2242)
\end{itemize}

\paragraph{BFFHQ.}
We created the dataset using the Flickr-Faces-HQ (FFHQ) Dataset~\cite{karras2019style}, which contains facial information, e.g., head pose and emotions. Out of these features, we selected age and gender as two attributes with a strong correlation. Our dataset includes 19200 images for training (19104 for bias-aligned and 96 for bias-conflicting) and 1000 samples for testing.

\subsection{Image Preprocessing}
Following~\cite{lee2021learning}, our model is trained and evaluated using images with the following fixed-size. For C-MNIST, the size is $28 \times 28$, while for C-CIFAR-10, it is $32 \times 32$. For BFFHQ, the size is $224 \times 224$. The images for C-CIFAR-10 and BFFHQ are preprocessed using random crop and horizontal flip transformations, as well as normalization along each channel (3, H, W) with the mean of (0.4914, 0.4822, 0.4465) and standard deviation of (0.2023, 0.1994, 0.2010). We do not use augmentation techniques to preprocess the images for C-MNIST.

\subsection{Performance Evaluation}
\label{subapp:performance}

\paragraph{Training Details.}
For training, we leverage the Adam~\cite{kingma2014adam} optimizer with default parameters (i.e., betas = (0.9, 0.999) and weight decay = 0.0) provided in the PyTorch\texttrademark framework. 

For learning rate, we define two different learning rates, $\texttt{LR}_{\texttt{DGW}}$ for our \texttt{DGW} modules, and $\texttt{LR}$ for the remaining modules in our method, including the encoders and classifiers. $\texttt{LR}$ for C-MNIST is set to 0.01. While $\texttt{LR}_{\texttt{DGW}}$ is 0.0005 for C-MNIST-2$\%$, 0.002 is for the remaining ratios of datasets for C-MNIST. For C-CIFAR-10, \texttt{LR} is 0.001, and $\texttt{LR}_{\texttt{DGW}}$ is 0.0001. For BFFHQ, $\texttt{LR}$ is set to 0.0001 whereas 0.0002 is for $\texttt{LR}_{\texttt{DGW}}$.

For each dataset, we utilize StepLR for learning rate scheduling. The decaying step is set to 10K for all datasets ($t_{\text{aug}}$ in Algorithm~\ref{alg:dgw_training_stragegies} in the main text), and the decay ratio is set to 0.5 for both C-MNIST and C-CIFAR-10 and 0.1 for BFFHQ. Referring to~\cite{lee2021learning}, we also leverage the scheduling for the learning rate after the feature augmentation is performed.

We define a set of hyperparameters $(\lambda_{\text{re}}, \lambda_{\text{swap}_{b}},  \lambda_{\text{swap}}, \lambda_{\text{ent}})$ for our proposed loss functions (Section~\ref{subsec:training_schemes} in the main text). $(10, 10, 1, 0.01)$ is set for the ratio of $0.5\%$ of C-MNIST, and $(15, 15, 1, 0.01)$ for the ratio of $1\%, 2\%$, and $5\%$ of C-MNIST. We set $(1, 1, 1, 0.01)$ for C-CIFAR-10, and $(2, 2, 0.1, 0.01)$ for BFFHQ.

Due to the architectural property of \texttt{ASA} module (Secion~\ref{subsec:asa_module} in the main text) in our approach, there are three hyperparameters, $C$ for the number of concepts, $d$ for the dimension of the latent concept embedding, and $T$ for the number of iterations to update the concept slots (Algorithm~\ref{alg:csa_module} in the main text). As described in the order of $(C, d, T)$, for C-MNIST, $(3, 8, 1)$ for the ratio of $0.5\%$, $(2, 8, 2)$ for $1\%$, $(5, 8, 1)$ for $2\%$, and $(2, 8, 3)$ for $5\%$. For C-CIFAR-10, $(6, 16, 3)$ for the ratio of $1\%$, and $(5, 16, 3)$ for the ratio of $0.5\%, 2\%$ and $5\%$. We use $(10, 32, 3)$ for BFFHQ.

In our proposed mixup strategy (Section~\ref{subsec:mixup} in the main text), there is the hyperparameter $\beta$ to choose the mixing coefficient $\alpha \sim \text{Beta}(\beta, \beta)$.
For BFFHQ, we set 0.5, whereas 0.2 for C-MNIST and C-CIFAR-10.

We conveniently provide the script files, including all hyperparameter setups above, in our Git repository (Section~\ref{app:reprod}) to reproduce our performance evaluation.


\subsection{Analysis for Attribute-Centric Representation Learning}
\label{subapp:attribute_centric}

\paragraph{Implementation details.}
In our research, we highlighted the importance of visualizing the mechanisms that enable our model to distinguish between intrinsic and biased attributes across a variety of datasets. Our Debiasing Global Workspace (DGW) includes both the \texttt{ASA} and \texttt{CA} modules, each generating attention masks. However, for the purposes of visualization, we specifically utilize the attention mask $\mathbf{A}^{\texttt{CA}}$ produced by the \texttt{CA} module. This choice stems from the dual QKV (Query, Key, Value) computation process around both modules. In our method, the \texttt{ASA} module updates the concept slots (with the slots acting as Q and the KV coming from the attribute embedding generated by the attribute encoder), followed by the \texttt{CA} module, which updates the attribute embedding (with the embedding serving as Q). As a result, the attention mask from the \texttt{CA} module provides especially insightful information regarding the focus areas of our model. 

Visualizing the \texttt{CA} module's attention mask deepens our understanding of how the model prioritizes and processes key features for decision-making. This enhanced visibility offers a more transparent and interpretable explanation of how our model navigates the problem of distinguishing between intrinsic and biased attributes across the datasets of C-MNIST, C-CIFAR-10, and BFFHQ. This section elaborates on our approach to visualize the attention masks within these specific datasets.

\paragraph{Initialization of Concept Slots.}
The initialization of the number of concept slots plays a critical role in our model. This initial setup is essential for tailoring the model's attention mechanisms to each dataset, enabling a nuanced understanding of how the model discerns diverse attributes. Accordingly, we designated the following initial values for the concept slots (denoted as $C$) across various datasets :
\begin{itemize}
    \item For C-MNIST, $C$ is set to 2, reflecting its relatively simple attribute composition.
    \item For C-CIFAR-10, $C$ is set to 10, accommodating its broader array of distinguishable features.
    \item For BFFHQ, $C$ is set to 10, ensuring the model's capability to capture a wide range of human facial features.
\end{itemize}
This deliberate initialization not only facilitates a deeper comprehension of the model's dynamics across different datasets but also underscores the importance of recognizing the diversity of attributes that are deemed significant within each specific context.

\paragraph{Visualization Thresholding.}
To visualize the most relevant areas identified by the \texttt{CA} module, we used a thresholding technique on the attention mask. This allows us to highlight only the regions of the image that the model deems essential for making predictions. The threshold values range from 0.0000 to 0.9999, enabling us to fine-tune the visualization based on the activation levels of the attention mask. By adjusting the threshold, we could exclude less relevant areas, focusing solely on the top activated regions. This selective visualization provides deeper insights into how the model distinguishes between intrinsic and biased attributes across different datasets. For each dataset, we systematically varied the threshold to visualize the attention mask's top activated areas. This approach ensures that only regions surpassing the specified threshold are displayed, offering a clear view of what the model considers crucial for its decision-making process. The visualization of these attention masks offers a transparent and interpretable view of the model's inner workings, highlighting its ability to focus on relevant features while ignoring potential biases.

\paragraph{Additional Visualization on C-MNIST dataset.}
\begin{figure*}[t!]
	\centering
	\includegraphics[width=0.75\linewidth]{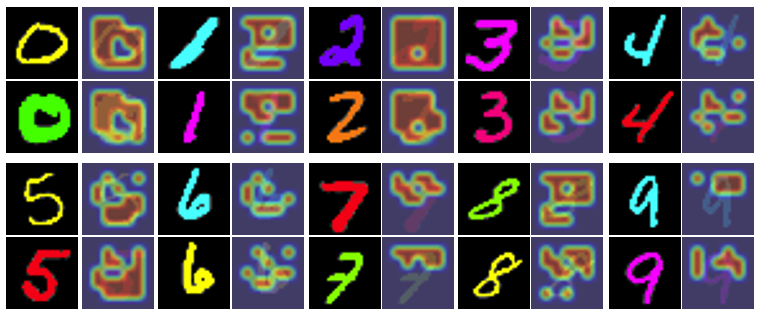}
	\caption{Visualization of attention masks $\mathbf{A}^{\texttt{CA}}_{i}$ generated by the intrinsic component $\texttt{DGW}^{i}$ for the C-MNIST dataset}\label{appfig:cmnist_intrinsic_full} 
\end{figure*}
\begin{figure*}[t!]
	\centering
	\includegraphics[width=0.5\linewidth]{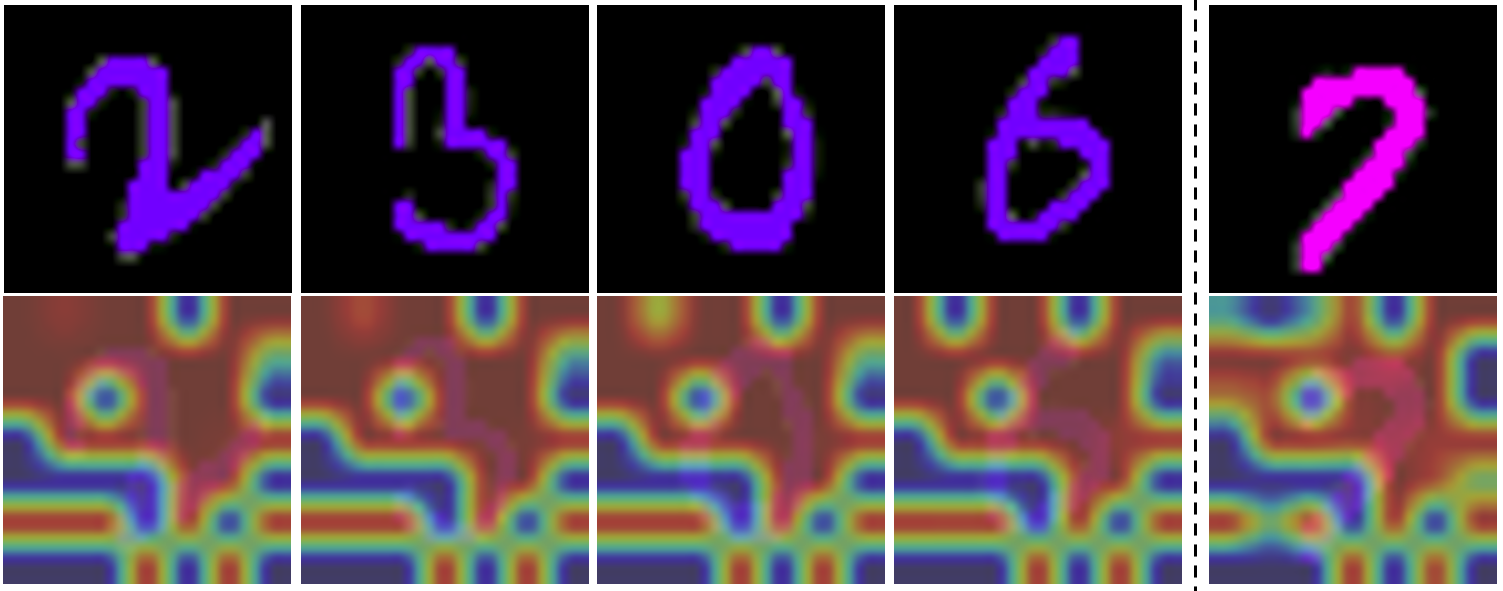}
	\caption{In this visualization from the C-MNIST dataset, attention masks $\mathbf{A}^{\texttt{CA}}_{b}$ generated by $\texttt{DGW}^{b}$ reveal patterns of recognition based on color. Digits displayed in the same color, such as 2, 3, 0, and 6, share the almost same attention mask pattern. Notably, the digit 7, presented in pink, is closely related to purple, showcasing how similar colors lead to similar attention mask pattern.}\label{appfig:cmnist_bias_sim} 
\end{figure*}
Fig. \ref{appfig:cmnist_intrinsic_full} displays the attention masks $\mathbf{A}^{\texttt{CA}}_{i}$ generated by the intrinsic module $\texttt{DGW}^{i}$ for the C-MNIST dataset, covering all digits from 0 to 9. These images show that our model focuses on the shape of the digits, ignoring their texture or color.

In comparison, Fig. \ref{appfig:cmnist_bias_sim} illustrates the attention masks $\mathbf{A}^{\texttt{CA}}_{b}$ generated by the bias component $\texttt{DGW}^{b}$ for the same dataset. The figure is key to understanding how the model responds to digits that appear in similar colors. When looking at the purple digits 2, 3, 0, and 6, it's clear that the masks are very similar. This similarity shows that our model pays more attention to the color of the digits when deciding how to react to them. When we look at the number 7, which is in pink, we see that its attention mask looks a lot like those of the purple digits. This comparison shows that our model is sensitive to small differences in color. It suggests that even slight color changes, like those between pink and purple, are enough for the model to react noticeably. These observations help us see how the DGW model can tell the difference between shape and color in what it learns from visual data. The model's ability to focus on the relevant features for each task can help us understand and manage the influence of different attributes in visual recognition.

\paragraph{Visualization on BFFHQ dataset.}
\begin{figure*}[t!]
	\centering
	\includegraphics[width=0.5\linewidth]{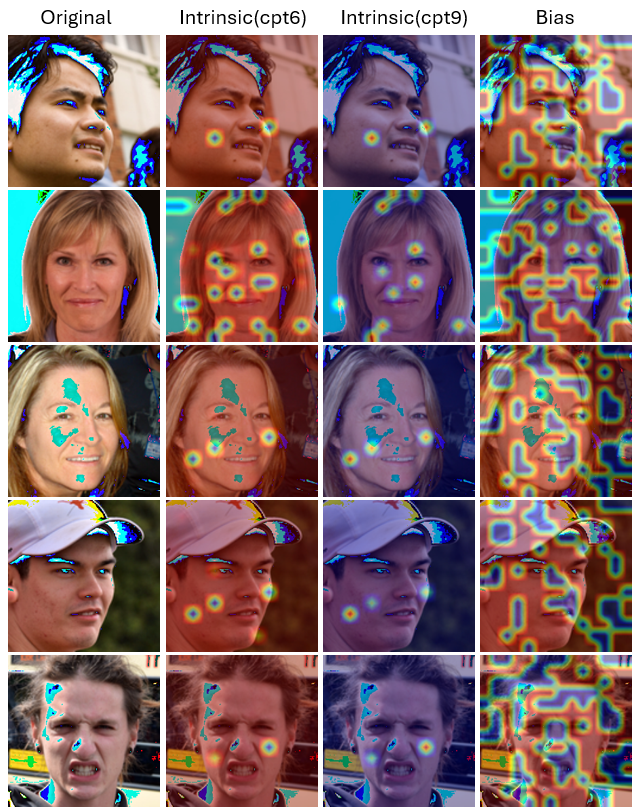}
	\caption{The figure showcases five sets of face images arranged in four columns. The first column presents the original image. The subsequent two columns display the attention mask $\mathbf{A}^{\texttt{CA}}_{i}$ interpreted by the intrinsic component $\texttt{DGW}^{i}$ at concept slots 6 and 9, respectively. The last column depicts the images with attention masks $\mathbf{A}^{\texttt{CA}}_{b}$, highlighting the regions associated with the bias component $\texttt{DGW}^{b}$. The intrinsic columns reveal focused activation on selective facial features, while the bias column shows widespread, less structured patterns across the faces.}
    \label{appfig:bffhq_mask} 
\end{figure*}
In contrast to the observations noted in Section~\ref{subsec:interpretable_analysis} in the main text, DGW exhibits a distinctive behavior when applied to the BFFHQ dataset. As depicted in Fig. \ref{appfig:bffhq_mask}, the DGW demonstrates complementary behavior not between intrinsic and bias components as seen in the C-CIFAR-10 dataset, but rather within the intrinsic components themselves. This novel interaction specifically manifests between the 6th and 9th concept embeddings within the intrinsic module $\texttt{DGW}^{i}$, labeled as `cpt6' and `cpt9', respectively. Specifically, a complementary relationship is evident between concept slots 6 and 9 within $\texttt{DGW}^{i}$, while the bias component $\texttt{DGW}^{b}$ displays what appears to be chaotic and geometric masks that are difficult to interpret by human.

This divergence in behavior is attributed to the unique characteristics of the BFFHQ dataset. Differing from the other datasets evaluated, BFFHQ presents a binary classification problem (male/female), but with an added complexity owing to its focus on human facial shapes. To accurately classify gender, our model seemingly concentrates on specific facial regions. For instance, cpt9 (the 9th concept embedding within $\texttt{DGW}^{i}$) highlights the cheeks area, suggesting its relevance in determining gender. Complementarily, cpt6 (the 6th concept embedding) directs attention to adjacent facial regions. This indicates a targeted approach where the model prioritizes certain facial features over others for decision-making, sidelining less pertinent areas as part of the bias component, as depicted in Fig. \ref{appfig:bffhq_mask}. Such a strategy underlines the model's adeptness at identifying and leveraging critical features for classification while efficiently filtering out noise or irrelevant data in scenarios that demand a nuanced distinction between classes.

\subsection{Quantitative and Qualitative Analysis}
\label{subapp:additional_analysis}

\paragraph{t-SNE and Clustering.}
\label{subsubapp:clustering}
\begin{figure*}[t!]
	\centering
	\includegraphics[width=0.95\linewidth]{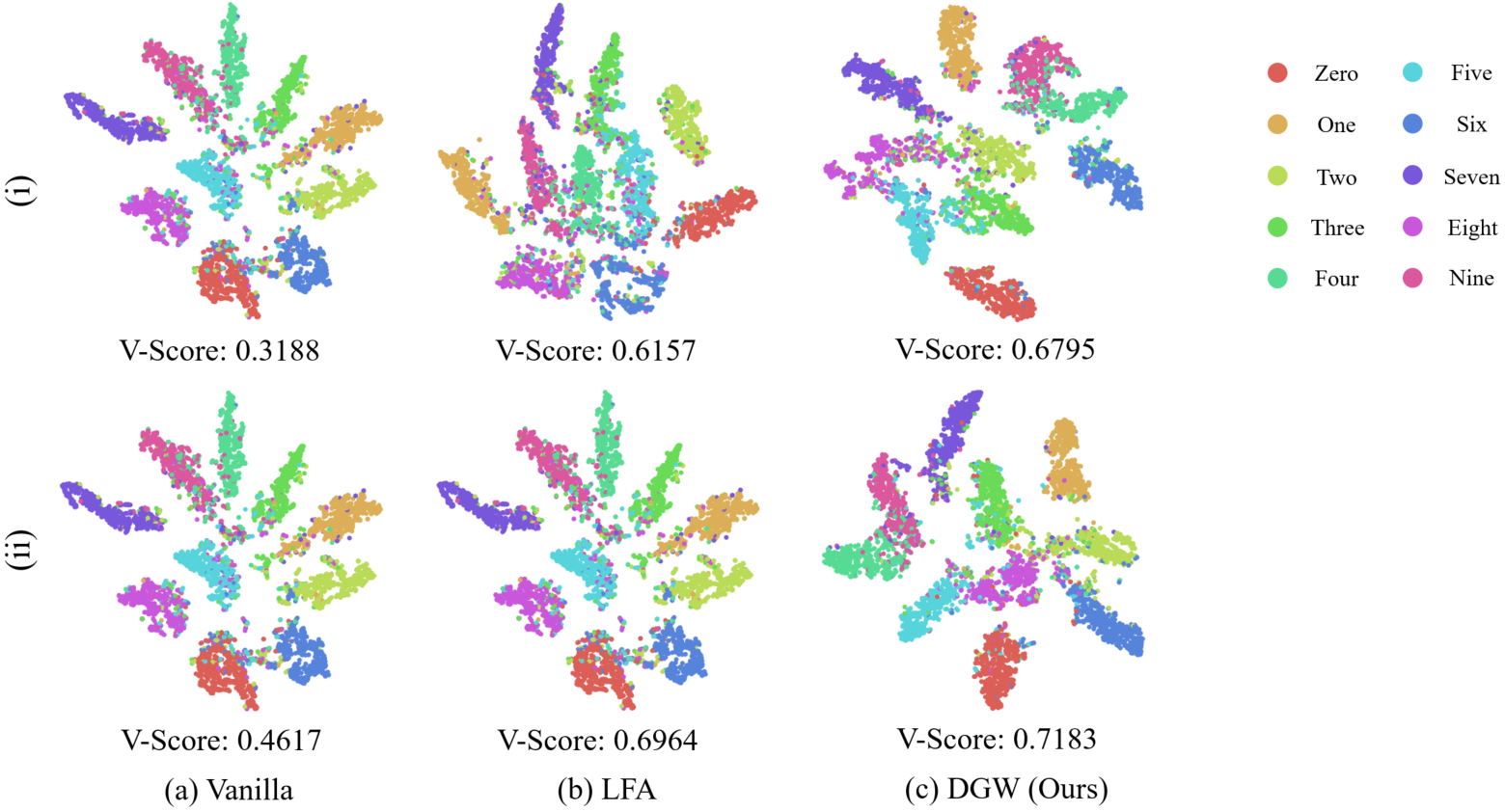}
	\caption{t-SNE plots for intrinsic features on C-MNIST (with (i) 1.0$\%$ and (ii) 2.0$\%$ settings).}
    \label{appfig:tsne_plot_1020} 
\end{figure*}
\begin{figure*}[t!]
	\centering
	\includegraphics[width=0.9\linewidth]{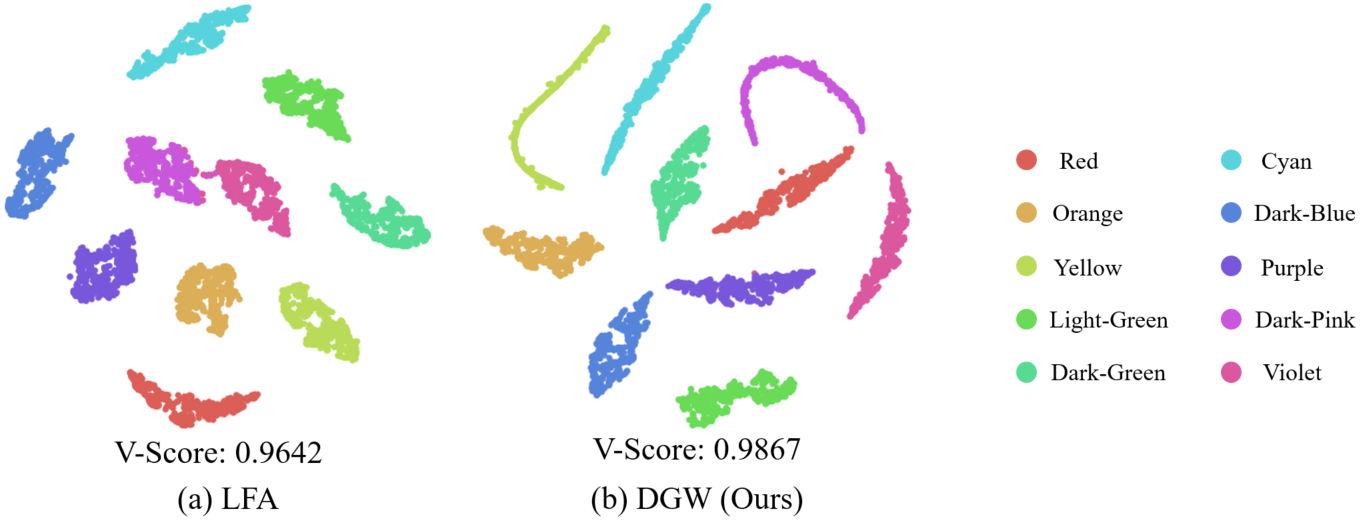}
	\caption{t-SNE plots for bias features on C-MNIST (with 0.5$\%$ setting).}\label{appfig:tsne_plot_cmnist05_bsupp} 
\end{figure*}
We provide more results with t-SNE plots and clustering scores with V-Score \cite{rosenberg2007v} as illustrated in Fig. \ref{appfig:tsne_plot_1020} and \ref{appfig:tsne_plot_cmnist05_bsupp}. V-Score has been widely used to evaluate clustering, a harmonic mean between homogeneity and completeness. For instance, a higher V-Score denotes tighter clusters for intra-class and better separation for inter-class, implying better homogeneity and completeness simultaneously.

In Fig. \ref{appfig:tsne_plot_1020}, intrinsic features from baselines and the intrinsic attribute encoder $\boldsymbol{\phi}^{i}$ of DGW are used. As shown in Fig. \ref{appfig:tsne_plot_1020}, in all cases, DGW shows a higher V-Score, implying the proposed method trains a model for better classification and capturing for intrinsic attributes compared to baselines. V-Scores of all methods of (ii) show better than (i) since more bias-conflicting samples are used for training in the (ii) setting than in the (i) setting.

In Fig. \ref{appfig:tsne_plot_cmnist05_bsupp}, features from bias attributes capturing layer of LFA and the bias attribute encoder $\boldsymbol{\phi}^{b}$ of DGW are utilized.
In this figure, compared to LFA, DGW shows a higher V-Score, whereas LFA generates more misclustered results. For both results of intrinsic and bias features, DGW outperforms baselines and shows consistent robustness. This represents that DGW enables the separation of intrinsic and bias attributes for improving the debiasing ability of a model.

\paragraph{Model Similarity.}
\label{subsubapp:similarity}
\begin{figure*}[t!]
	\centering
	\includegraphics[width=0.99\linewidth]{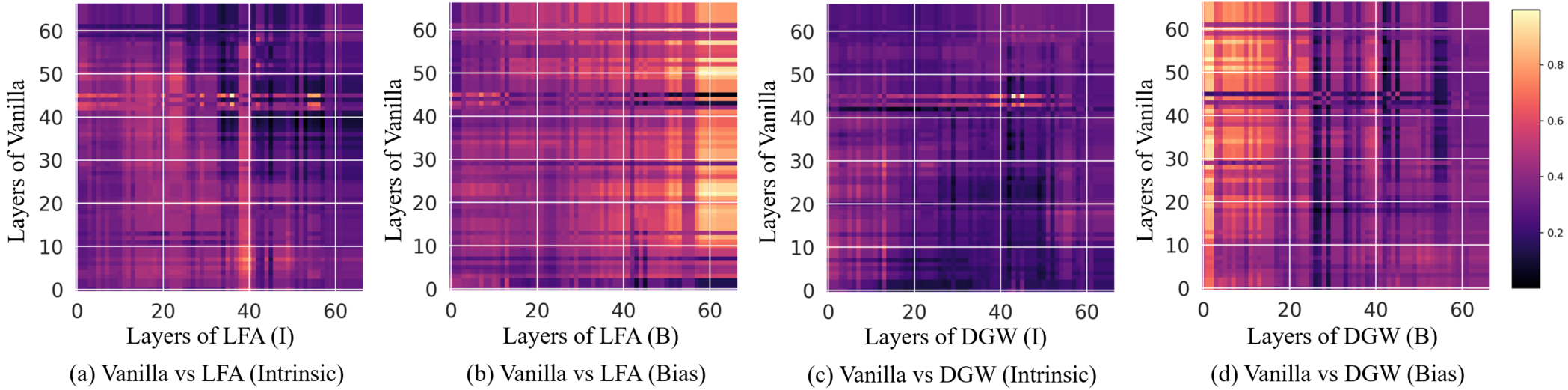}
	\caption{Representations of similarities for vanilla model and different methods with all pairs of layers on C-CIFAR-10 (5.0$\%$ setting). High similarity score denotes high values. }\label{appfig:model_sim_cifar50} 
\end{figure*}
\begin{figure*}[t!]
	\centering
	\includegraphics[width=0.99\linewidth]{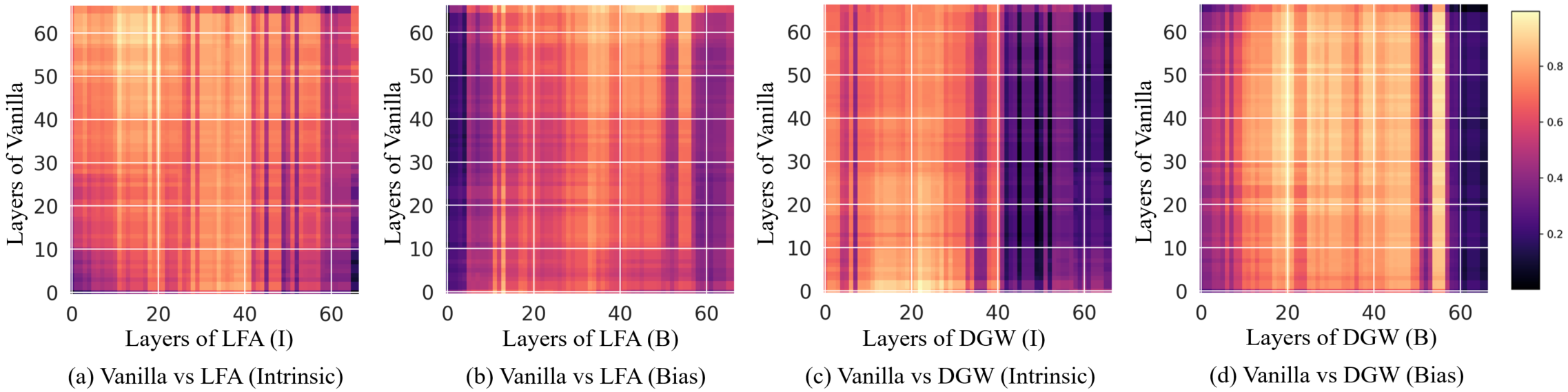}
	\caption{Representations of similarities for vanilla model and different methods with all pairs of layers on BFFHQ (0.5$\%$ setting). High similarity score denotes high values. }\label{appfig:model_sim_bffhq05} 
\end{figure*}
We plot model similarity with Centered Kernel Alignment (CKA)~\cite{raghu2021vision, kornblith2019similarity, cortes2012algorithms} to visualize the similarities between all pairs of layers with different models, which helps understand the behavior of models. 
The bias and intrinsic attribute encoders $\boldsymbol{\phi}^{b}$ and $\boldsymbol{\phi}^{i}$ in our approach are used for comparison.
As illustrated in Fig. \ref{appfig:model_sim_cifar50} and Fig. \ref{appfig:model_sim_bffhq05}, Vanilla and LFA possess similar weights in many layers regardless of intrinsic and bias attributes, represented with bright colors. 

In Fig. \ref{appfig:model_sim_cifar50}(b), the deeper layers show more similarities than the initial layers.
On the other hand, in Fig. \ref{appfig:model_sim_cifar50}(c) and (d), Vanilla and DGW show significantly lower similarity values compared to the LFA results. As shown in Fig. \ref{appfig:model_sim_cifar50}(d), there is contrast on columns, and there are not many highlighted colors on diagonal points, representing that DGW layers possess different weights, compared to Vanilla. This tendency is also different from LFA. 

In Fig. \ref{appfig:model_sim_bffhq05}, DGW results show less similarities compared to LFA for both intrinsic and bias cases.
This presents that DGW and baselines have different behaviors across layers of a model. Also, DGW affects the deeper layers more than the initial layers, where the deeper layer is the place where the attention module is inserted.
Therefore, the behavior of models trained by DGW differs from other methods across datasets and settings.

\paragraph{Model Reliability.}
\label{subsubapp:reliability}
To evaluate the generalizability of models, we measure Expected Calibration Error (ECE) and Negative Log Likelihood~(NLL) \cite{guo2017calibration}, where ECE is to measure calibration error and NLL is to calculate the probabilistic quality of a model. In detail, ECE aims to evaluate whether the predictions of a model are reliable and accurate, which is a simple yet sufficient metric for assessing model calibration and reflecting model generalizability \cite{guo2017calibration}.

In Table \ref{apptab:full_reliability_table}, in overall cases, DGW shows the lowest ECE, which represents better calibration and reliability of models. For C-MNIST, DGW presents a larger NLL compared to baselines. Since C-MNIST includes color bias only for the training set, DGW leads a model to prevent over-fitting while being less affected by bias, which eventually helps obtain a better model. This tendency, seen with DGW, is consistent across settings on C-MNIST. This observation can provide more insights to analyze and explain the characteristics of bias types and complexity in the dataset.
\begin{table}[t!]
\caption{ECE ($\%$) and NLL under different settings on C-MNIST and C-CIFAR-10.}
\begin{center}
\scalebox{0.84}{\begin{tabular}{l |cccccccc |cccccccc}
\toprule
Dataset & \multicolumn{8}{c|}{C-MNIST} & \multicolumn{8}{c}{C-CIFAR-10} \\ \midrule
Ratio ($\%$) & \multicolumn{2}{c}{0.5} & \multicolumn{2}{c}{1.0}  & \multicolumn{2}{c}{2.0} & \multicolumn{2}{c|}{5.0} & \multicolumn{2}{c}{0.5} & \multicolumn{2}{c}{1.0}  & \multicolumn{2}{c}{2.0} & \multicolumn{2}{c}{5.0} \\ 
& ECE & NLL & ECE & NLL & ECE & NLL & ECE & NLL & ECE &  NLL & ECE & NLL & ECE & NLL & ECE & NLL  \\ \midrule
Vanilla & 10.9 & \textbf{13.17}& 7.97 & \textbf{6.45} & 5.70 & \textbf{5.71} & 9.54 & 4.10 & 13.75 & 5.99 & 13.14 & 9.87 & 12.25 & 6.65 & 13.76 & 5.99 \\
LFA & 4.35 & 67.72 & 2.79 & 36.46 & 2.09 & 18.35 & 7.59 & \textbf{3.09} & 12.09 & 5.81& \textbf{11.45} & 7.27 & 10.25 & 5.14 & 7.56 & 3.09 \\
DGW & \textbf{3.41} & 271.71  & \textbf{2.03} & 143.36 & \textbf{1.73} & 41.44 & \textbf{1.61} & 20.19 &\textbf{11.85} & \textbf{5.71} & 11.53 & \textbf{6.88} & \textbf{9.96} & \textbf{4.41} & \textbf{7.55} & \textbf{3.01} \\
\bottomrule
\end{tabular}
}
\label{apptab:full_reliability_table}
\end{center}
\end{table}

\section{Limitations}
\label{app:limit}
Compared to existing approaches, our method is a novel debiasing method with attribute-centric information bottlenecks that enables learning latent attribute embedding to contribute to better performance and improve interpretability by visualizing how the model focuses on intrinsic and bias attributes differently. However, in this section, we explain some limitations of our proposed approach. 

We acknowledge that introducing our modules can increase training complexity, including model size and training time. This can be a trade-off between performance and transparency of decision-making, so further analysis is necessary. 

Furthermore, in Section~\ref{subapp:background}, despite our approach being inspired by Global Workspace Theory (GWT), its implementation is not ideally along with the theoretical ground of GWT because we leveraged two separate global workspaces ($\texttt{DGW}^{i}$ and $\texttt{DGW}^{b}$ in Section~\ref{subsec:training_schemes}) for our training convenience. We tried to test a single \texttt{DGW} module theoretically closer to the GWT than ours but witnessed the instability of performances on the biased datasets. To overcome this, we may adopt a specific masking technique in the single \texttt{ASA} module to manually specify the regions of slots for intrinsic and biased attributes separately.

\end{document}